\documentclass[10pt]{article} 
\usepackage[preprint]{tmlr}

\usepackage{amsmath,amsfonts,bm}

\def\eqref#1{equation~\ref{#1}}

\def\1{\bm{1}}

\DeclareMathAlphabet{\mathsfit}{\encodingdefault}{\sfdefault}{m}{sl}
\SetMathAlphabet{\mathsfit}{bold}{\encodingdefault}{\sfdefault}{bx}{n}

\usepackage{hyperref}
\usepackage{url}
\usepackage{booktabs}       
\usepackage{amsfonts}       
\usepackage{subcaption}
\usepackage{tcolorbox}
\usepackage{amsmath}
\usepackage{amssymb}
\usepackage{mathtools}
\usepackage{amsthm}
\usepackage{nicefrac}       
\usepackage{microtype}      
\usepackage{xcolor}         
\usepackage{microtype}
\usepackage{graphicx}
\usepackage{wrapfig}

\title{Understanding the Staged Dynamics of Transformers in Learning Latent Structure}

\author{%
  \name Rohan Saha \email rsaha@ualberta.ca \\
  \addr Department of Computing Science\\
  University of Alberta\\
  Edmonton, Canada
  \AND
  \name Farzane Aminmansour \email aminmans@ualberta.ca \\
  \addr Department of Computing Science \\
  University of Alberta\\
  Edmonton, Canada
  \AND
  \name Alona Fyshe \email alona@ualberta.ca \\
  \addr Department of Computing Science and Psychology\\
  University of Alberta\\
  Edmonton, Canada
}

\def\month{MM}  
\def\year{YYYY} 
\def\openreview{\url{https://openreview.net/forum?id=XXXX}} 

\begin{document}

\maketitle

\begin{abstract}
  Language models have proven to be capable of learning the intricate latent structures in language. However, the dynamics of how such models acquire the components of that structure remain poorly understood, leaving them open to the criticism that models just remix training data. In this work, we use the Alchemy benchmark as a controlled setting in which we investigate how models acquire the different latent structure components of the data.
  We train a small decoder-only transformer on three controlled task variants: 1) inferring missing steps from partial contextual information, 2) composing simple rules to solve multi-step sequences, and 3) decomposing complex multi-step examples to infer intermediate states. By factorizing each task into interpretable components, we show that the model learns the different latent structure components in discrete \textit{stages} and in a \textit{coarse-to-fine} manner. Using inference-time ablations, we demonstrate that when certain latent structures are masked or replaced, the model can use an alternative intermediate latent structure to solve the task. We also observe an asymmetry: the model composes single step transitions robustly, but struggles to decompose complex examples to discover the atomic steps. Our findings provide insight into \textit{how} a transformer model acquires latent structure, offering a detailed view of how capabilities evolve during training.
\end{abstract}


\section{Introduction}
\label{sec:introduction}

In the context of language modeling, it is clear that transformers can learn latent structure. For example, models learn syntactic hierarchies without supervision, with attention heads encoding structures such as dependency trees \citep{hewittStructuralProbeFinding2019, manningEmergentLinguisticStructure2020, ravishankarAttentionCanReflect2021}, and attention mechanisms encoding causal graphs \citep{nichaniHowTransformersLearn2024}. However, due to the complexity of latent structures in language, it is difficult to study the progression of learning latent structure, and even more difficult to show that models are truly generalizing rather than just remixing training data \citep{bender2021dangers}.
\par
Here, we explore a setting with a clear and controlled latent structure: the Alchemy benchmark~\citep{wang2021alchemy}. The latent structures in Alchemy allow us to study the learning progression in multiple settings by carefully controlling what the model observes during training. 
Across our many settings, we observe staged learning dynamics: performance exhibits plateaus, followed by sudden jumps, suggesting that models acquire capabilities in stages rather than continuously. Such observations have been linked to several factors, including hierarchical structure in data, hidden progress during learning, and staged learning in transformers and language models \citep{abbeStaircasePropertyHow2021, barakHiddenProgressDeep2022, edelmanEvolutionStatisticalInduction2024,singhWhatNeedsGo2024,chenSuddenDropsLoss2025,songOutofdistributionGeneralizationComposition2025,gopalaniWhatHappensLoss2025}. However, the complexity of these settings makes it difficult to attribute the stages to what is actually learned. Using the Alchemy benchmark provides us with a controlled setting to study the learning dynamics and attribute the observed stages to specific aspects of Alchemy’s inherent latent structure.

\par

\begin{figure*}[!htb]
    \centering
    \includegraphics[width=\textwidth]{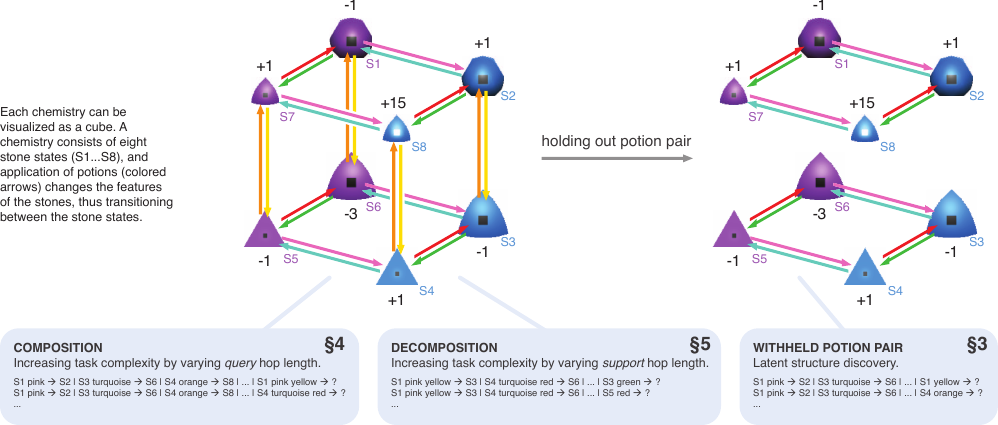}
    \caption{Overview of Alchemy chemistry structure and experimental tasks. A chemistry graph (left panel) contains eight stone states (vertices) connected by potion-induced transitions (edges), that change stone features. Right callout: experiment to investigate latent structure learning dynamics. For a chemistry, transitions for a randomly selected potion pair are withheld (e.g., yellow/orange); the model needs to correctly infer their effects. To solve this task, the model needs to correctly align the two disconnected halves created from withholding the potion pair (right panel). Left callout: composition task where the model needs to solve a multi-hop query given all 1-hop (single-step) support transitions. Middle callout: decomposition requires solving a 1-hop query given all multi-hop transitions in the support , given a hop length. In our experiments, we fully enumerate the stone features in the support and query. The part after the last $|$ is the query. Numbers in the callouts denote the respective sections of the paper discussing results. Figure adapted from \citet{wang2021alchemy}.}
    \label{fig:example_of_a_chemistry_graph_composition_decomposition}
\end{figure*}

In the Alchemy setting, players start with a stone whose properties can be altered by applying various potions, changing the stone's features accordingly. The set of stones and the transitions between them, induced by potions, form a ``chemistry''. Each chemistry is represented as a directed graph with eight vertices arranged in a cubic structure (Figure \ref{fig:example_of_a_chemistry_graph_composition_decomposition}, left panel). 
We investigate latent structure learning dynamics by withholding observations of latent features, and varying task complexity. These controlled experiments demonstrate that rapid performance improvements align with the successful learning of the different aspects of the chemistry. 

We conduct depth-first analyses of latent structure learning in a setting with cubic latent structure. We train a small decoder-only transformer model ($\sim$ 2M parameters) \citep{vaswaniAttentionAllYou2017} on three formulations of the Alchemy task (Figure \ref{fig:example_of_a_chemistry_graph_composition_decomposition}). 1) We strategically withhold information about the effects of specific potions and test the model's ability to infer the effects of these missing potions (Figure \ref{fig:example_of_a_chemistry_graph_composition_decomposition}, right callout).
2) We investigate compositionality by providing the model with all the single potion effects and test its ability to compose these single potions to solve novel, multi-step potion sequences (Figure \ref{fig:example_of_a_chemistry_graph_composition_decomposition}, left callout). 3) We test the model's decomposition ability by providing it with multi-step potion sequences and requiring it to infer the single-step potion effects (Figure \ref{fig:example_of_a_chemistry_graph_composition_decomposition}, middle callout).
\par

Our central contribution is to analyze the staged dynamics of latent structure learning in a controlled setting, allowing us to attribute the observed stages to specific latent structure components, which the model learns in a coarse-to-fine manner. We factorize the model’s overall accuracy into interpretable components corresponding to distinct latent properties, and show that the observed plateaus and jumps are not random but correspond to acquiring specific interpretable sub-skills. Furthermore, through ablation studies, we show that a transformer model can leverage alternative intermediate latent structures to solve the task, but only when certain information is masked or replaced.
Our analyses and results complement prior work on staged learning by identifying what is learned at each stage, and what intermediate latent structure a transformer model uses to learn the task.
We also conducted layer freezing experiments to provide supporting evidence for the coarse-to-fine latent structure acquisition, and identify layer specific plasticity windows during which freezing interventions delay subsequent stage acquisition.
Concretely:

\begin{enumerate}
    \item We show that transformers learn latent structure in a coarse-to-fine manner via distinct stages: the model first narrows the set of possible outcomes, then learns the exact target. We make this progression explicit by factorizing overall accuracy into interpretable sub-skills, showing that the stages correspond to learning different latent structure components (Section \ref{sec:latent_structure_learning}).

    \item We demonstrate that, when information about a latent structure is masked or replaced, the model can leverage alternative information about the latent structure to complete the task (Section \ref{sec:reward_adjacency_token_ablation}).

    \item We show that, under composition, there is no increase in convergence time of the coarse-to-fine learning stages. 
    When composing single-step potion transitions to solve multi-step potion sequences, the model learns the stages at approximately the same time during training, across varying numbers of steps in the potion sequence (Section \ref{sec:composition_experiment}).
    
    \item We identify an asymmetry between composition and decomposition in latent structure learning. When all the support examples are provided, model convergence is delayed with increasingly long multi-step potion sequences to infer a single-step potion transition. The convergence delay is reflected in the delayed learning of later stages (Section \ref{sec:intermediate_stone_state_inference_experiment}).

\end{enumerate}



The accompanying \href{https://anonymous.4open.science/r/alchemy_latent_structure-C24E/README.md}{code} and the data generation pipeline are released to encourage further research on understanding the transformer dynamics on latent structure learning.


\section{Methods}
\label{sec:methods}
In this section, we introduce the Alchemy dataset, its latent structures, and properties. We then introduce notation, followed by the transformer model architecture and the training details.

\subsection{Dataset}
\label{sec:dataset}
The Alchemy dataset \citep{wang2021alchemy} is a compositional meta-reinforcement learning benchmark designed to test a model's ability to discover and exploit latent structures.
In Alchemy, players apply potions to a stone, changing its state. Interactions between the stones and potions are defined by a chemistry, and chemistries differ between rounds. All chemistries share a latent cubic graph structure that describes the interactions (Figure \ref{fig:example_of_a_chemistry_graph_composition_decomposition}, left panel). \par

To solve Alchemy, the model cannot memorize the chemistry specific transitions between the stones, and instead must learn to discover and exploit the invariant latent structures that are consistent across all chemistries.
While the original alchemy dataset contains chemistries with missing edges (incomplete graphs), we only consider complete chemistries,\footnote{
Since incomplete chemistries are subgraphs of complete ones, we exclude them to ensure strict disjointness between train
and validation sets.} where each vertex has three outgoing edges.

The chemistries have the following invariances:
\begin{itemize}
    \item Each chemistry graph has 8 possible stone states positioned on the vertices of the cubic structure. Each stone is defined by four features: color, size, roundness, and reward. The first three perceptual features (`color', `size', `roundness') have three possible values.  The `color' feature can take values pink, violet, blue, `size' can take values small, medium, large, and `roundness' can take values pointy, medium\_round, and round.
    The `reward' feature can take one of four values (-3, -1, +1, +15). This creates $3 \times 3 \times 3 \times 4 = 108$ unique possible stones across all chemistries, and a chemistry contains exactly 8 stones. 
    \item Potions come in fixed complementary pairs (red/green, yellow/orange, pink/turquoise) that are consistent across all chemistries. Applying a potion changes the stone's features, moving the stone state as dictated by an edge connecting two vertices, and complementary potions always have inverse effects. For any given chemistry, each potion color corresponds to a single latent transition rule that can be applied to a vertex. Crucially, a potion's transition rule corresponding to one axis may not align with a single perceptual feature, and thus a single potion can cause multiple perceptual features to change at once. For example, in the chemistry shown in Figure \ref{fig:example_of_a_chemistry_graph_composition_decomposition} (left panel), the yellow potions adjust the roundness and size by a single level, but keep the color unchanged. By contrast, the pink potion leaves the stone's size and roundness unchanged, but shifts its color by two levels (pink $\rightarrow$ violet $\rightarrow$ blue).
    \item The reward feature follows a structured distribution consistent across all chemistries: +15 is adjacent only to +1, -3 is adjacent only to -1, +1 is adjacent to +15 and -1, and -1 is adjacent to +1 and -3. Thus, applying a potion changes the reward feature based on the stone's position in the chemistry, with complementary potions inducing inverse reward transitions. Unlike in reinforcement learning settings, where the objective is to maximize reward, here the reward is simply a stone feature.
\end{itemize}

\subsection{Notation and Task Formulation}
We frame the Alchemy task as a supervised learning problem. The model is provided with a set of \textbf{support} (contextual) examples from a specific chemistry, using which the model must respond to a \textbf{query} example. We formalize these notions below.

\par
Given a complete chemistry graph, the support $S$ consists of triplets with $M$ example transitions:
$S = \{(x_{s_m}, z_{s_m}, y_{s_m})\}_{m=1}^M$.
Each support example represents a transformation in the chemistry, where 
$x_{s_m}$ is the \textit{start stone state}, 
$z_{s_m}$ is the \textit{sequence of potions applied}, 
and $y_{s_m}$ is the resulting \textit{end stone state}.
We randomize the order of the examples in $S$ to prevent any ordering biases.
Let $Q = \{q_i\}_{i=1}^{N}$ denote the set of all possible queries for a chemistry, where each $q_i = (x_i, z_i)$. We ensure that $S$ and $Q$ are disjoint. 
We train the model to learn $p(y_i \mid q_i, S)$, i.e., given a query $q_i$, the model must predict the correct output stone $y_i$, using the support $S$ to infer the latent chemistry.

\par
To systematically vary task complexity, we introduce `hop length' (${hl}$): the length of the potion sequence $z$. For instance, a 2-hop example $(hl = 2$) corresponds to applying two potions in succession moving from the start stone state to the end stone state in the chemistry (Figure \ref{fig:example_of_a_chemistry_graph_composition_decomposition}). We vary $hl$ for both support ($hl_{support}$) and query ($hl_{query}$).  
\par

Using this task formulation, we design three systematic experiments (callouts in Figure \ref{fig:example_of_a_chemistry_graph_composition_decomposition}). (1) \textbf{Withheld potion pair}: we vary the content of $S$ with $hl_{support}=1$ but withhold all examples for a randomly selected potion pair. The query $q_i$ with $hl_{query}=1$ then tests the model's ability to infer the effect of the withheld potions (right callout). 
(2) \textbf{Composition}: we test the model's ability to compose 1-hop support transitions ($hl_{support}=1$) to solve multi-hop queries ($hl_{query} \in \{2,3,4,5\}$) (left callout). (3) \textbf{Decomposition}: we test the model's ability to decompose complex multi-hop support examples ($hl_{support} \in \{2,3,4,5\}$) to solve a 1-hop query ($hl_{query}=1$) (middle callout).

\par
\subsection{Training Details}
\label{sec:model}

Following prior work on learning dynamics \citep{gopalaniWhatHappensLoss2025}, we restrict our analysis to a small-scale ($\sim$2M parameter) causal transformer \citep{vaswaniAttentionAllYou2017} with 4 hidden layers, where each layer has 4 attention heads. We use an embedding dimension of 256, a feedforward dimension of 512, and a dropout of 0.1. The output is a classification layer over all the 108 possible stones states (Section \ref{sec:dataset}).

\par
Based on work in learning from contextual examples \citep{garg_what_can_transformers_learn, chanDataDistributionalProperties2022, singhTransientNatureEmergent2023, lakeHumanlikeSystematicGeneralization2023}, we concatenate all support examples $S$ followed by the query; the number of support examples and the resulting input length vary across tasks and hop lengths (Appendix \ref{appendix:training_details_hyperparameter}). We use standard cross-entropy loss on a 108-way classification head over the target stone state from the final sequence position. We use a 90 / 10 train - validation split over the 1,571 complete chemistries, ensuring that validation chemistries are unseen during training. We train all models for 1000 epochs, with a batch size of 32, and with three random initializations (seeds). We use the earliest converging runs for each seed to analyze all the learned stages, but show that failure runs also exhibit staged dynamics (Appendix \ref{appendix:hyperparameter_sensitivity}).

\paragraph{Tokenization details:} We used a task specific custom vocabulary to tokenize the input. We serialized the stone states by enumerating their four features as separate tokens: color, size, roundness, and the reward feature. Each support example is organized as: input stone $x$ (4 tokens) followed by the potion sequence $z$ (one token per potion), an input-output separator \texttt{<io>}, the output stone $y$ (4 tokens), and an item separator \texttt{<item\_sep>}. The query consists of the input stone followed by the potion(s). 
The following is an instance of a support example where two potion transitions are applied to a stone:

\texttt{pink large round -1  green yellow <io> pink medium pointy -1 <item\_sep>}
\par
A query example can similarly be tokenized as follows (the query does not have tokens \texttt{<io>} or \texttt{<item\_sep>}):

\texttt{blue large round +1 yellow}

\par
Since we only use complete chemistries, all sequences for a task / hop length combination are of the same length; no padding is required and no truncation is applied. We also controlled for the number of training samples across hop lengths for each task. We summarize the support sizes and input lengths for all the tasks and the associated hop lengths in Table \ref{tab:appendix_data_input_lengths} in Appendix \ref{appendix:training_details_hyperparameter}.


\section{Learning the Withheld Potion Pair}
\label{sec:latent_structure_learning}


In this experiment, the model must infer the effect of a withheld potion pair using the remaining support transitions. Intuitively, withholding transitions for a potion pair creates two \textit{disconnected} halves of the chemistry graph (Figure \ref{fig:example_of_a_chemistry_graph_composition_decomposition}, right panel) that must be aligned to correctly answer the query. This contrasts previous work where models may rely on surface-level patterns or counting statistics \citep{mittalDoesLearningRight2024}, and instead requires learning the underlying transition rules.

\subsection{Analysis of withheld potion learning stages}
\label{sec:analytical_study_4_edge_held_out}
\begin{figure*}[!htb]
    \centering
    \begin{subfigure}[t]{0.49\linewidth}
        \centering
        \includegraphics[width=1.0\linewidth]{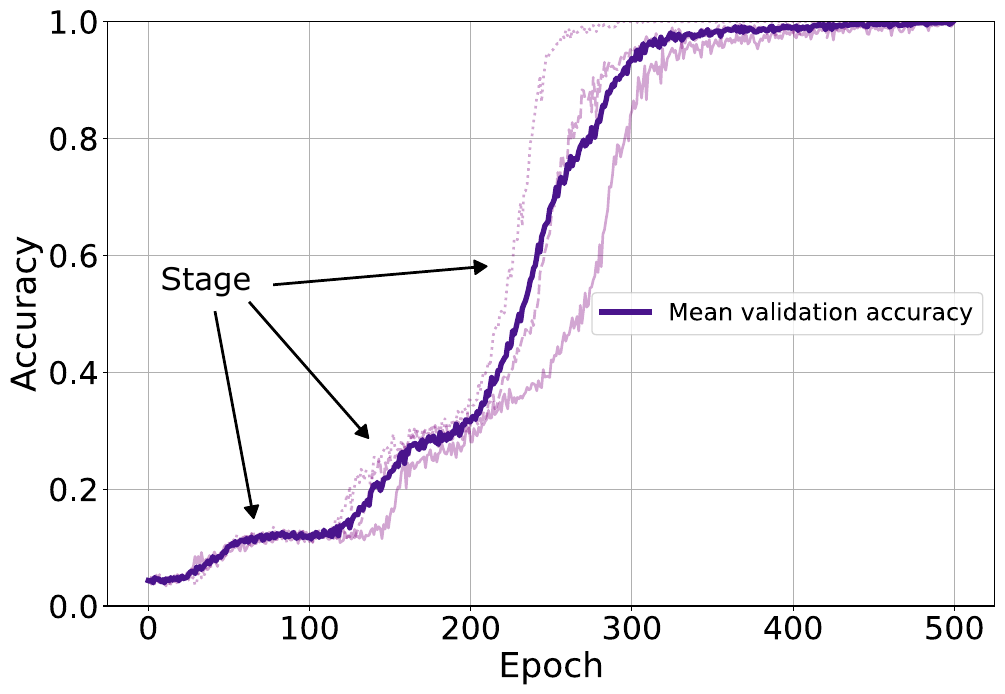}
        \caption{Model accuracy exhibits stages via rises \& plateaus.}
        \label{fig:4_way_edge_completion_performance}
    \end{subfigure}
    \hfill
    \begin{subfigure}[t]{0.49\linewidth}
        \centering
        \includegraphics[width=1.0\linewidth]{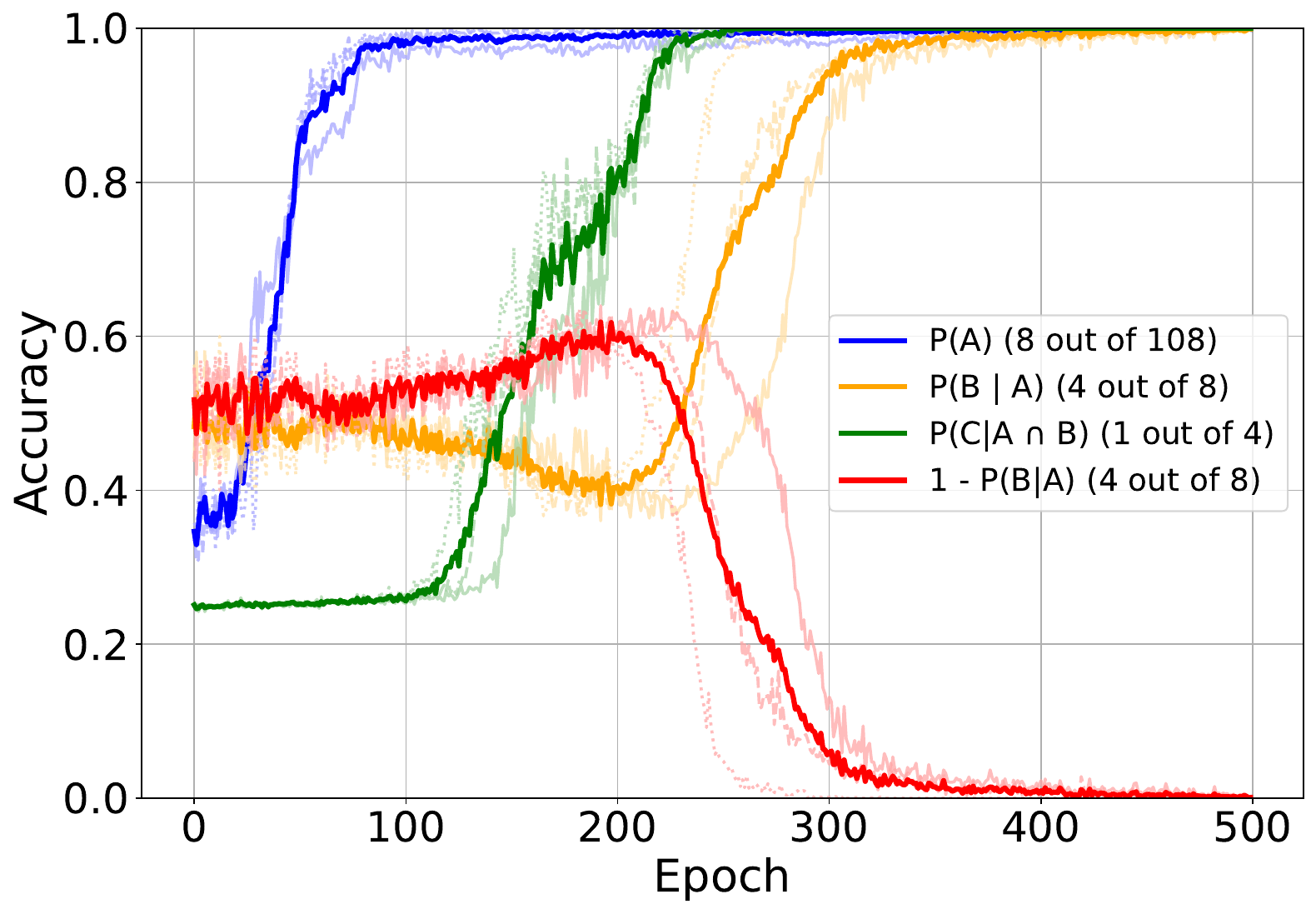}
        \caption{Task factorization reveals staged learning dynamics.}
        \label{fig:4_edge_held_out_stages}
    \end{subfigure}
    \caption{
    Staged learning of latent structure components for a withheld potion pair (500 epochs). (a) Overall model accuracy showing discrete stages indicated by rises and plateaus. (b) Factorization into components: $\mathbb{P}[A]$ (blue, in-support), $\mathbb{P}[B | A]$ (orange, correct half), and $\mathbb{P}[C | A \cap B]$ (green, exact match). Red: incorrect half accuracy. $\mathbb{P}[A]\;\mathbb{P}[B\,|\,A]\;\mathbb{P}[C\,|\,A\cap B]$ recovers the overall accuracy in Figure \ref{fig:4_way_edge_completion_performance}. Lighter and darker curves denote individual runs and the mean respectively.}
    \label{fig:4_edge_held_out_phasic_learning}
\end{figure*}

Figure \ref{fig:4_way_edge_completion_performance} shows the results, which depicts multiple stages in the accuracy curve. We systematically study these stages by factorizing the task into interpretable events associated with the task's latent structures. \par

\textbf{Sets and Events:}

Consider the chemistry $chem$ in Figure \ref{fig:example_of_a_chemistry_graph_composition_decomposition}, where withholding one potion pair creates two disconnected halves. Let $T_{chem}$ be the set of eight stones in the support $S$ containing the remaining transitions. We denote the half containing the query stone $x_i$ as $H_{chem}^{(same)}$, and the half containing the ground truth $y_i$ as $H_{chem}^{(other)}$. Solving the task requires narrowing predictions along these constraints. Given $chem$, query $q_i$, and prediction $\hat y_i$, we define:

\[
\begin{aligned}
A &:= \{\hat y_i \in T_{chem}\} &&\text{(in-support)},\\
B &:= \{\hat y_i \in H_{chem}^{(other)}\} &&\text{(correct half)},\\
C &:= \{\hat y_i = y_i\} &&\text{(exact match)}.
\end{aligned}
\]
These events leverage meaningful structural properties of the task and represent \textit{sub-skills} that the model must achieve to solve the task. For instance, $A$ is a meaningful event because the correct answer $(y_i)$ is always in $T_{chem}$, reflecting whether the model has learned to restrict its predictions from the global space of 108 stones to the current chemistry. $B$ is a meaningful event as withholding transitions for a potion pair creates two disconnected halves, and applying the withheld transition moves stones between halves, so identifying the correct half is a necessary step to learn the task.

By the chain rule, the overall accuracy in Figure \ref{fig:4_way_edge_completion_performance} is the product of the conditional probabilities: $\mathbb{P}[C] \;=\; \mathbb{P}[A]\;\mathbb{P}[B\,|\,A]\;\mathbb{P}[C\,|\,A\cap B]$.
Note that according to the definition of $T_{chem}$ and $H_{chem}^{(other)}$, when $\hat y_i = y_i$, $C \subset B \subset A$ always holds. We plot these components in Figure \ref{fig:4_edge_held_out_stages}.
For completeness, we also show $\mathbb{P}[A\cap B^c\,|\,A]=1-\mathbb{P}[B\,|\,A]$ (red) - predicting the same (incorrect) half $H_{chem}^{(same)}$. 
We define a \textit{stage} as a training interval characterized by a rise and plateau in the overall model accuracy (Figure \ref{fig:4_way_edge_completion_performance}) and use the factorized component accuracy to analyze each stage. 
The stages in Figure \ref{fig:4_edge_held_out_stages} are as follows:
\begin{enumerate}
  \item \emph{Stage 1: In-support (epochs $\approx$20–100)}. $\mathbb{P}[A]$ (blue) quickly rises achieving perfect accuracy, indicating that the model quickly restricts its predictions to the valid support set. 

  \item \emph{Stage 2: Exact match given the correct half (epochs $\approx$100–240).}
  $\mathbb{P}[C\,|\,A\cap B]$ (green) quickly rises indicating that the model can identify the target for $q_i$ given the correct half ($\mathbb{P} [B \mid A]$). Yet, the model has not yet learned $\mathbb{P}[B \mid A]$ (at $\sim$ 50\% chance accuracy). 

  \item \emph{Stage 3: Correct half given in-support (epochs $\approx$ 220–320).}
  The model completes learning $\mathbb{P}[B | A]$ (orange) resulting in the complete learning of the task. 

\end{enumerate}

Figure \ref{fig:4_edge_held_out_stages} shows counterintuitive learning dynamics: the exact match accuracy ($\mathbb{P}[C \mid A \cap B]$) rises before the correct half accuracy ($\mathbb{P} [B \mid A]$). This effect is caused by an adjacency bias, where the model exploits the reward structure to identify the set of stones adjacent to the query start stone $x_i$. For example, in Figure \ref{fig:example_of_a_chemistry_graph_composition_decomposition} (left), if $x_i = S8$ (reward +15), its adjacent stones (via a 1-hop transition) always have a +1 reward feature (in both same and other halves). This structure allows the model to identify adjacent neighbors without having access to the full latent graph.


\par


The transient reduction in correct half performance ($H_{chem}^{(other)}$ - orange curve; epochs $\approx 100-200$ in Figure \ref{fig:4_edge_held_out_stages}) confirms this effect. Crucially, two out of three adjacent stones reside in the same (incorrect) half, whereas the true target is in the correct half. Therefore, as the model learns to predict adjacent stones, it becomes biased towards the incorrect half, causing the observed correct half accuracy dip. Finally, in stage 3, the model predicts the correct half and learns the task. Appendix \ref{appendix:reward_adjacency_grouped_by_reward} provides a detailed analysis of the adjacency bias. To investigate whether the model learned the correct half structure (event $B$), we conducted numerous inference-time ablation experiments, which we discuss next.

\subsection{Causal Corroboration of Adjacency Bias via Inference-Time Ablations}
\label{sec:reward_adjacency_token_ablation}
To provide additional evidence that the counterintuitive stage ordering (Section \ref{sec:latent_structure_learning}) is due to reward adjacency bias, we conducted \textbf{three inference-time ablation experiments on two levels}.
We used the trained models from Section \ref{sec:latent_structure_learning}, and only conducted ablations at test time at every epoch. Specifically, we ablated the `reward feature' token of the input during inference on each model checkpoint. The ablations are:
\begin{enumerate}
    \item[I.] Token Replacement: replace the `reward' feature token(s) with an unknown <unk> token from the model vocabulary.
    \item[II.] Zero-Ablation:  replace the `reward' feature token(s) with an embedding vector of all zeroes.
    \item[III.] Attention Masking:  modify the attention mask across all layers to block information propagation from `reward' token positions.
\end{enumerate}

The levels of analyses are:

\begin{itemize}
    \item Global: We applied the ablations on all the `reward' tokens in the model input (support and query).
    \item Local: We applied the ablations only on the `reward' token of the query.
\end{itemize}
\par

\begin{figure}[!htb]
    \centering
    \includegraphics[width=0.55\linewidth]{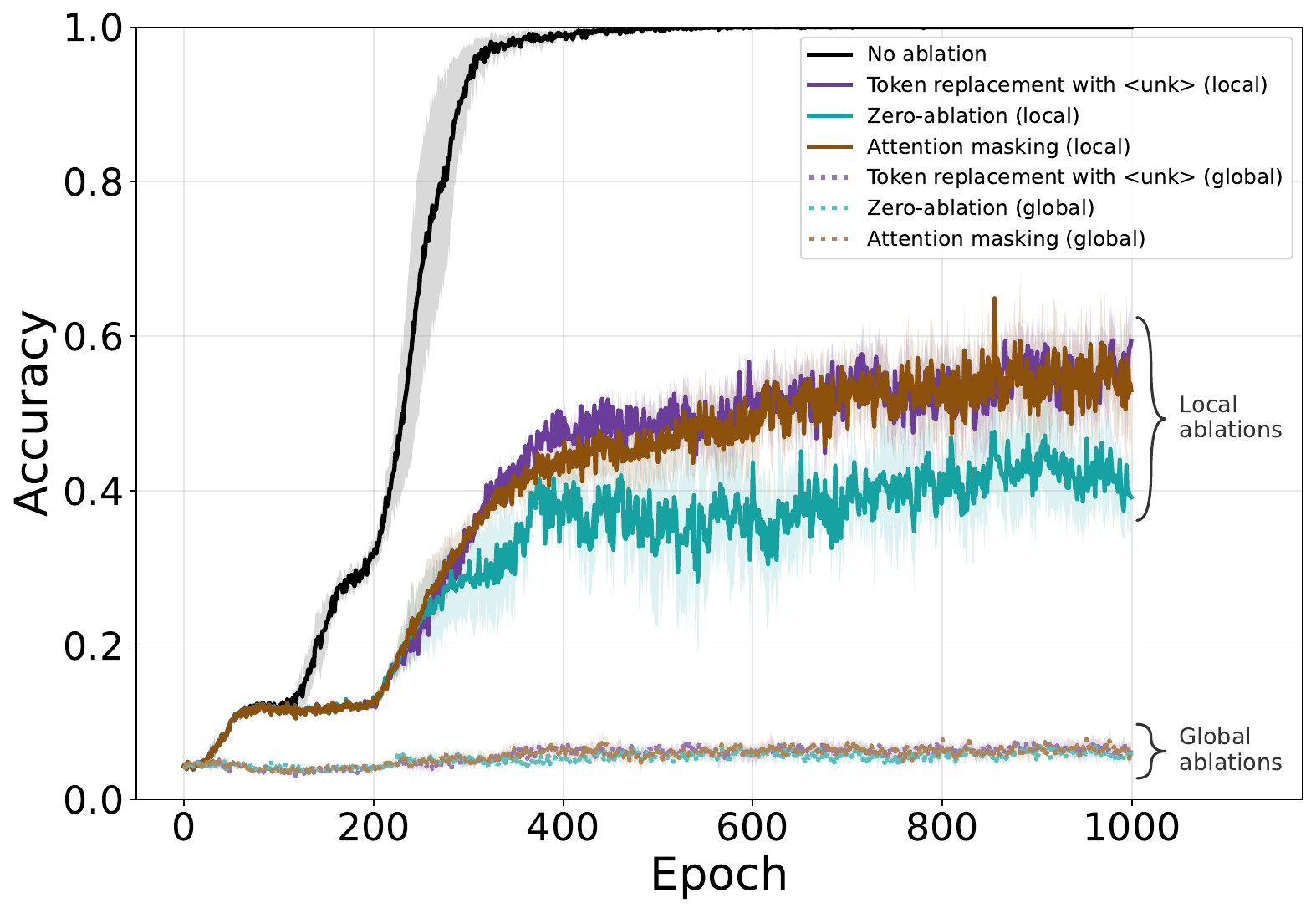}
    \caption{Global and local `reward' feature token ablation results. The black curve denotes the baseline run (no ablation - same as Figure \ref{fig:4_way_edge_completion_performance}). When ablating all the reward feature values in the support and the query (global ablation), model accuracy collapses (<10\%) indicating that the reward is necessary for the model to solve the task. When ablating only the query reward (local ablation), the model reaches accuracy >40\% but <100\%. Dark lines show the mean accuracy and the error bars denote the standard error of the mean.}
    \label{fig:withheld_potion_reward_ablation_all_results}
\end{figure}

\begin{figure*}[!htb]
    \centering
    \begin{subfigure}[t]{0.32\linewidth}
        \centering
        \includegraphics[width=\textwidth]{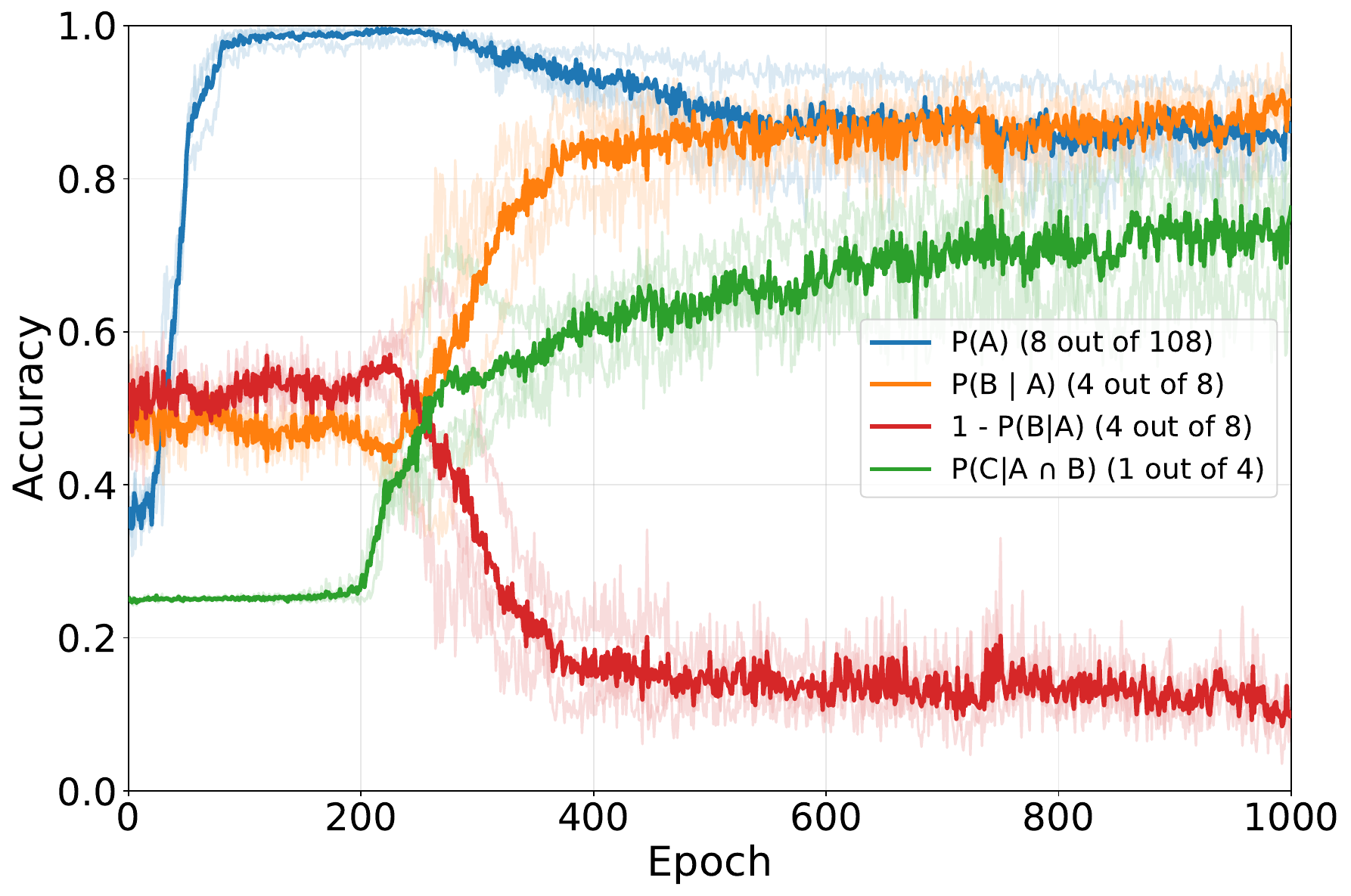}
        \caption{<unk> token replacement.} 
        \label{fig:replace_query_reward_with_unk}
    \end{subfigure}
    \begin{subfigure}[t]{0.32\linewidth}
        \centering
        \includegraphics[width=\textwidth]{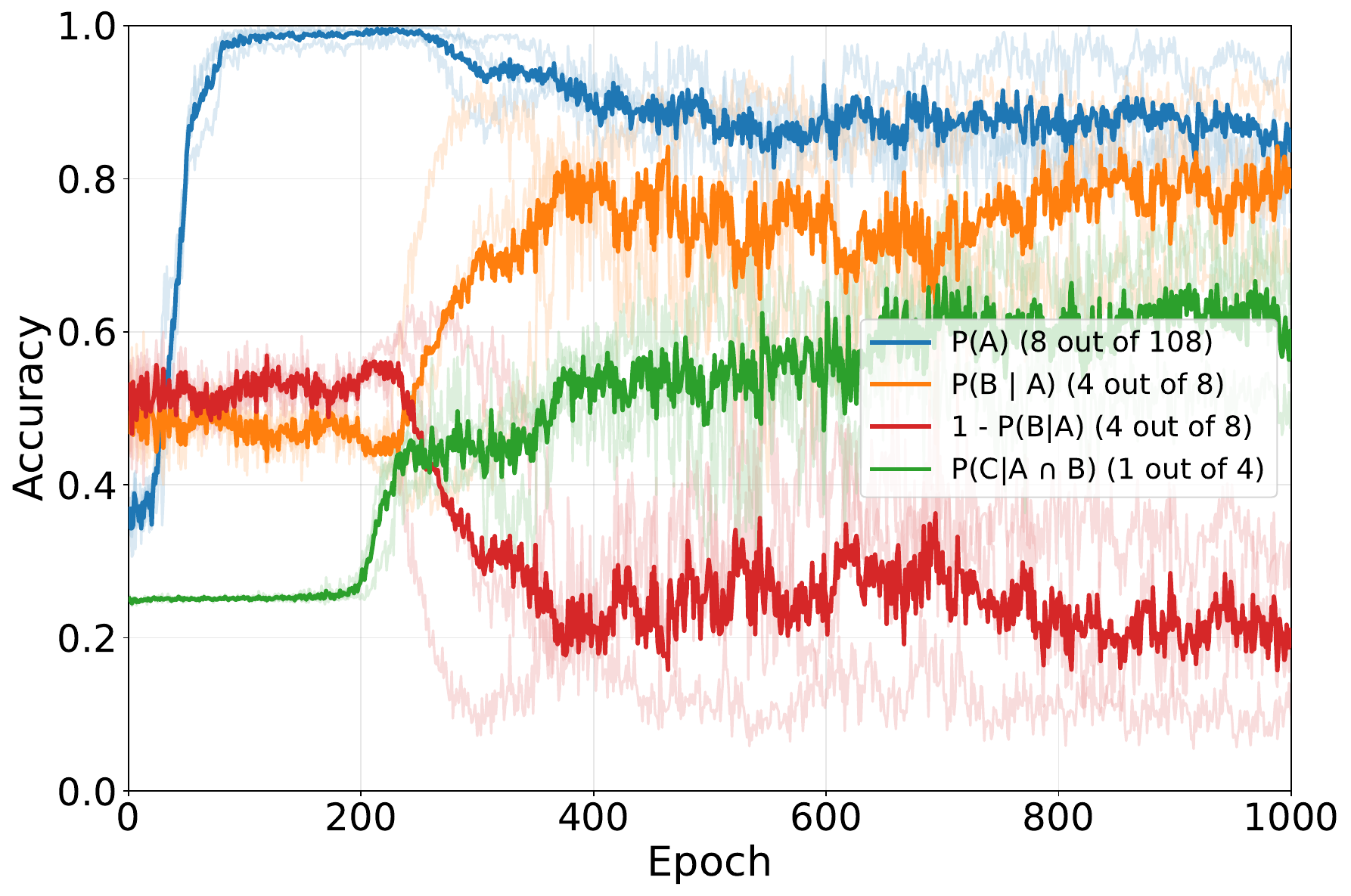}
        \caption{Query reward zero-ablation.}
        \label{fig:replace_query_reward_with_zero}
    \end{subfigure}
    \begin{subfigure}[t]{0.32\linewidth}
        \centering
        \includegraphics[width=\textwidth]{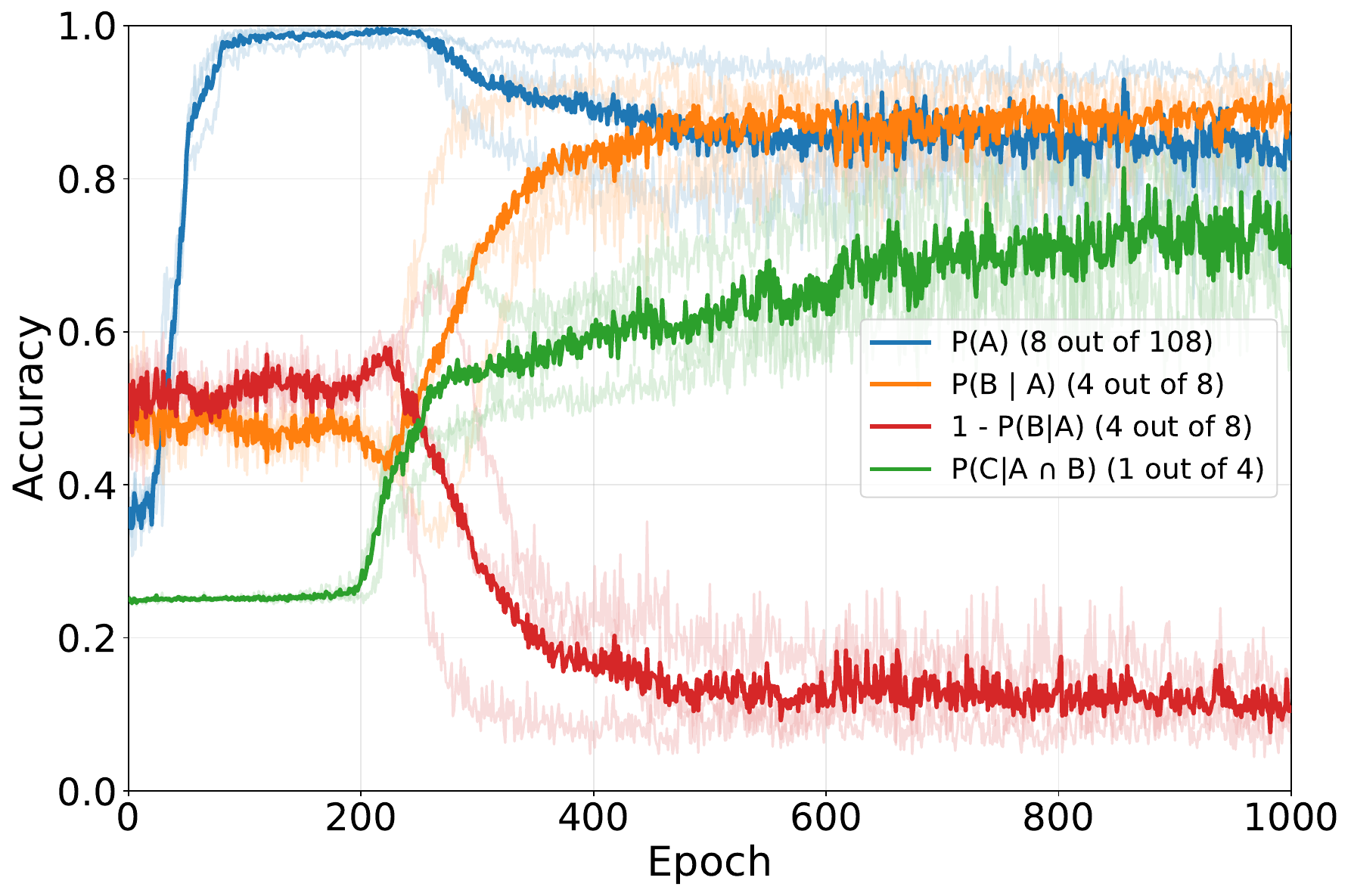}
        \caption{Attention masking ablation.}
        \label{fig:mask_query_reward_at_attention}
    \end{subfigure}
    \caption{Staged dynamics results for the query stone reward feature (local) ablations for the withheld potion pair experiment. We observe that different local ablations flip the ordering of the $\mathbb{P}[C|A \cap B]$ and $\mathbb{P}[B|A]$ stages. (a): Staged dynamics when replacing the query stone reward feature value with <unk> token (token replacement). (b): Staged dynamics when replacing the query stone reward feature value with zero embedding (zero-ablation). (c) Staged dynamics when attention masking is applied to the query stone reward feature value (attention masking). All methods show that when the reward feature is ablated, the model uses the \textit{correct half} intermediate latent structure as an ``alternative strategy'' to restrict its predictions before learning the correct target, and there is no observed dip in correct half accuracy due to over-leveraging the reward latent structure.}
    \label{fig:held_out_query_reward_ablations}
\end{figure*}

Taken together, we have six total ablations; we show the overall accuracy results for all ablations (global and local) and the baseline (no ablation) in Figure \ref{fig:withheld_potion_reward_ablation_all_results}, and the stages for local ablations in Figure \ref{fig:held_out_query_reward_ablations}. 

\textbf{Effect of global ablations}: Applying ablations (I, II, III) to all reward feature tokens caused accuracy collapse (<10\%) across all methods, confirming the model's reliance on the reward features to solve the task.
\par
\textbf{Effect of local ablations}: Applying ablations only to the query reward allowed the model to reach higher accuracy (>40\%), but less than 100\%. The overall accuracy curves for the local ablation results exhibit stages denoted by rises and plateaus. 
The in-support stage ($\mathbb{P}[A]$) was still learned earliest, but the learning order of stages 2 and 3 was reversed compared to Figure \ref{fig:4_edge_held_out_stages} --- $\mathbb{P}[B|A]$ was consistently higher than $\mathbb{P}[C | A \cap B]$. 


\par
These ablation results suggest that during inference, when the query reward feature is available, the transformer model uses the reward-based adjacency latent structure as the intermediate stage before learning the correct target. When the query reward feature is ablated, the model uses the ``alternative strategy'' (correct-half structure) as the intermediate stage before learning the correct target.
In other words, ablating the query reward feature during inference at each checkpoint inhibits the model's ability to leverage the reward adjacency information, forcing the model to rely on the structural constraint of the two disconnected halves (Section \ref{sec:latent_structure_learning}). This results in the ordering where $\mathbb{P}[B|A]$ is consistently higher than $\mathbb{P}[C|A \cap B]$ (Figure \ref{fig:held_out_query_reward_ablations}).
\par
Our results are consistent with the counterintuitive stage ordering (Section \ref{sec:latent_structure_learning}) being driven by the reward adjacency bias. More importantly, the ablation results and the reordering of stages 2 and 3 reinforce the notion of the coarse-to-fine staged learning. In other words, the model consistently learns the broader coarse constraints (correct-half or reward-adjacency) before refining its predictions to the correct target. All in all, we show that the transformer model learns and can leverage the alternative strategy at different points during training, which was previously primarily studied in fully trained models \citep{wang2022interpretabilitywildcircuitindirect}.





\section{Convergence Invariance in Composition}
\label{sec:composition_experiment}
\begin{figure}[!htb]
    \centering
    \includegraphics[width=0.5\linewidth]{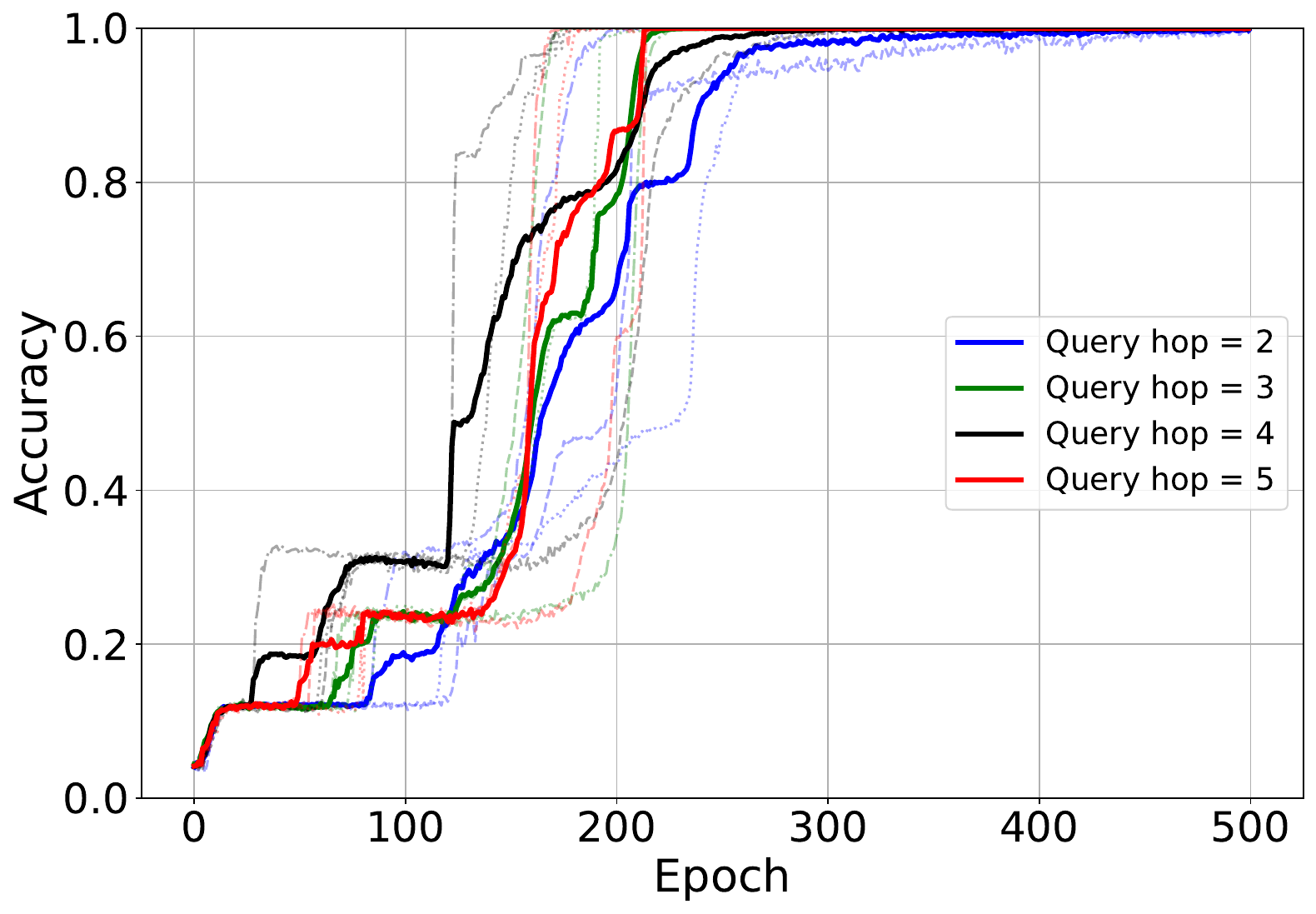}
    \caption{Overall accuracy for the composition task.  We observed no noticeable increase in convergence time for $hl_{query} \in \{2,3,4,5\}$. All hops reach convergence within 500 epochs. Lighter and darker curves show individual runs and the mean respectively.}
    \label{fig:composition_performance_all_hops}
\end{figure}

For the composition task, the model must solve multi-hop queries ($hl_{query} \in \{2,3,4,5\}$) given the support set $S$ of all possible 1-hop transitions ($hl_{support} = 1$) in the Alchemy cubic setting. In our experimental setup, we exclude examples with cycles (revisiting a stone in a sequence) and those with the same start and end stones.
Figure \ref{fig:composition_performance_all_hops} shows the staged learning dynamics for the composition task. We observed no increase in convergence time with varying query hop length, at least for $hl_{query} \in \{2,3,4,5\}$. 
This pattern is mirrored by similar stage timing across hops. 

\subsection{Analysis of composition learning stages}
\label{sec:analytical_study_of_composition}
\begin{figure*}[!htb]
    \centering
    \begin{subfigure}[t]{0.24\linewidth}
        \centering
        \includegraphics[width=\linewidth]{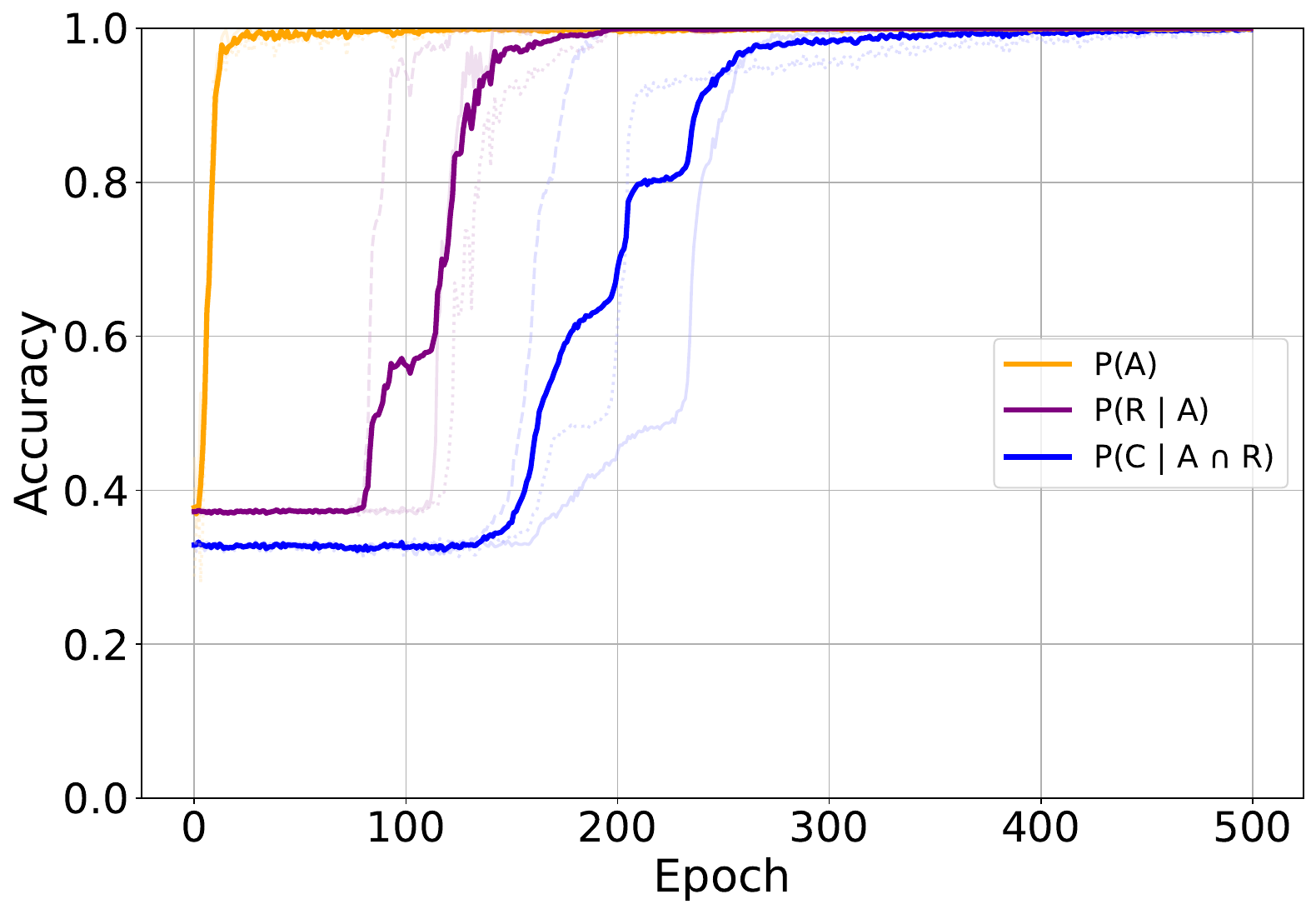}
        \caption{$hl_{query}=2$}
        \label{fig:composition_2_hop_phasic_learning}
    \end{subfigure}
    \hfill
    \begin{subfigure}[t]{0.24\linewidth}
        \centering
        \includegraphics[width=\linewidth]{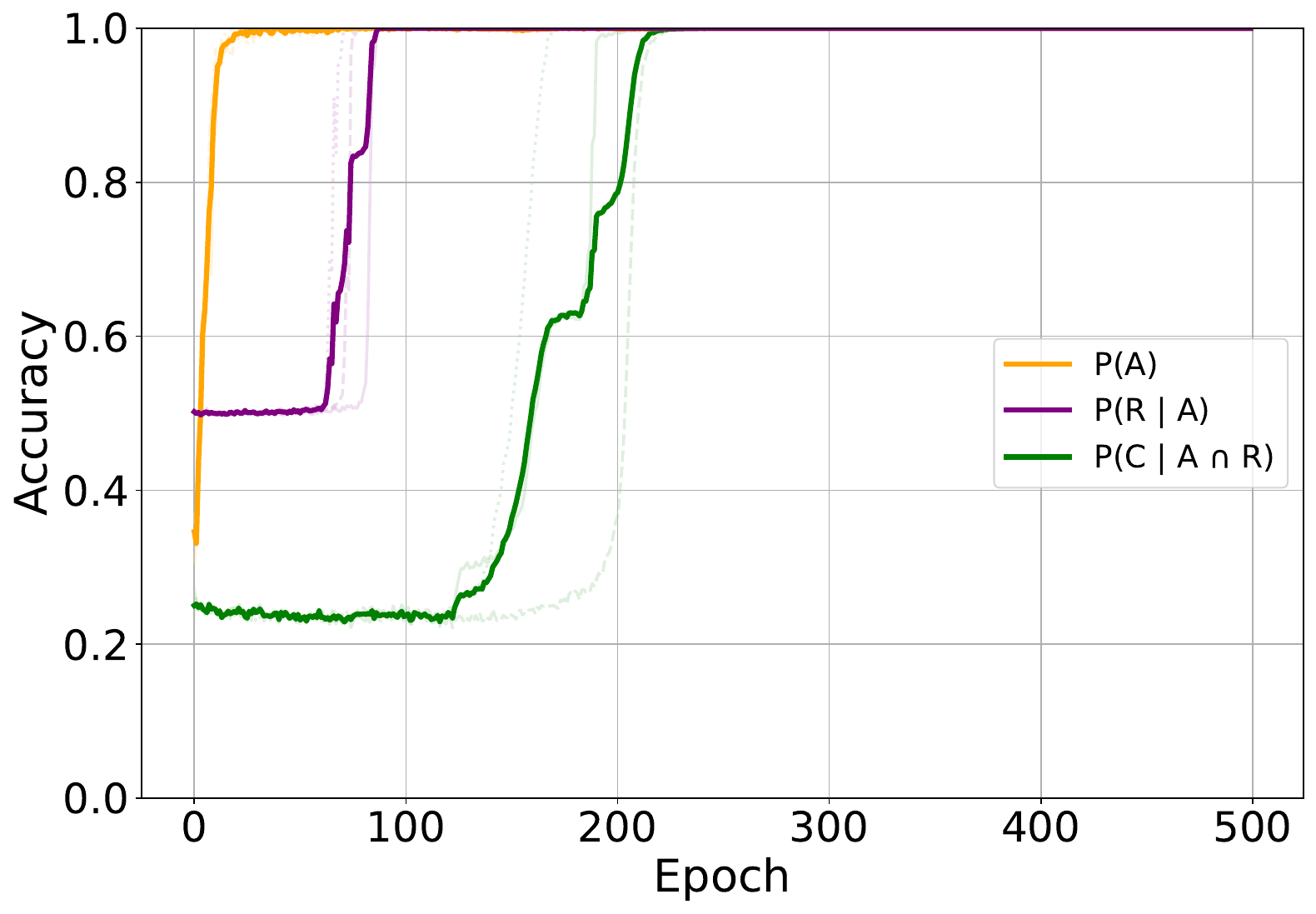}
        \caption{$hl_{query}=3$}
        \label{fig:composition_3_hop_phasic_learning}
    \end{subfigure}
    \hfill
    \begin{subfigure}[t]{0.24\linewidth}
        \centering
        \includegraphics[width=\linewidth]{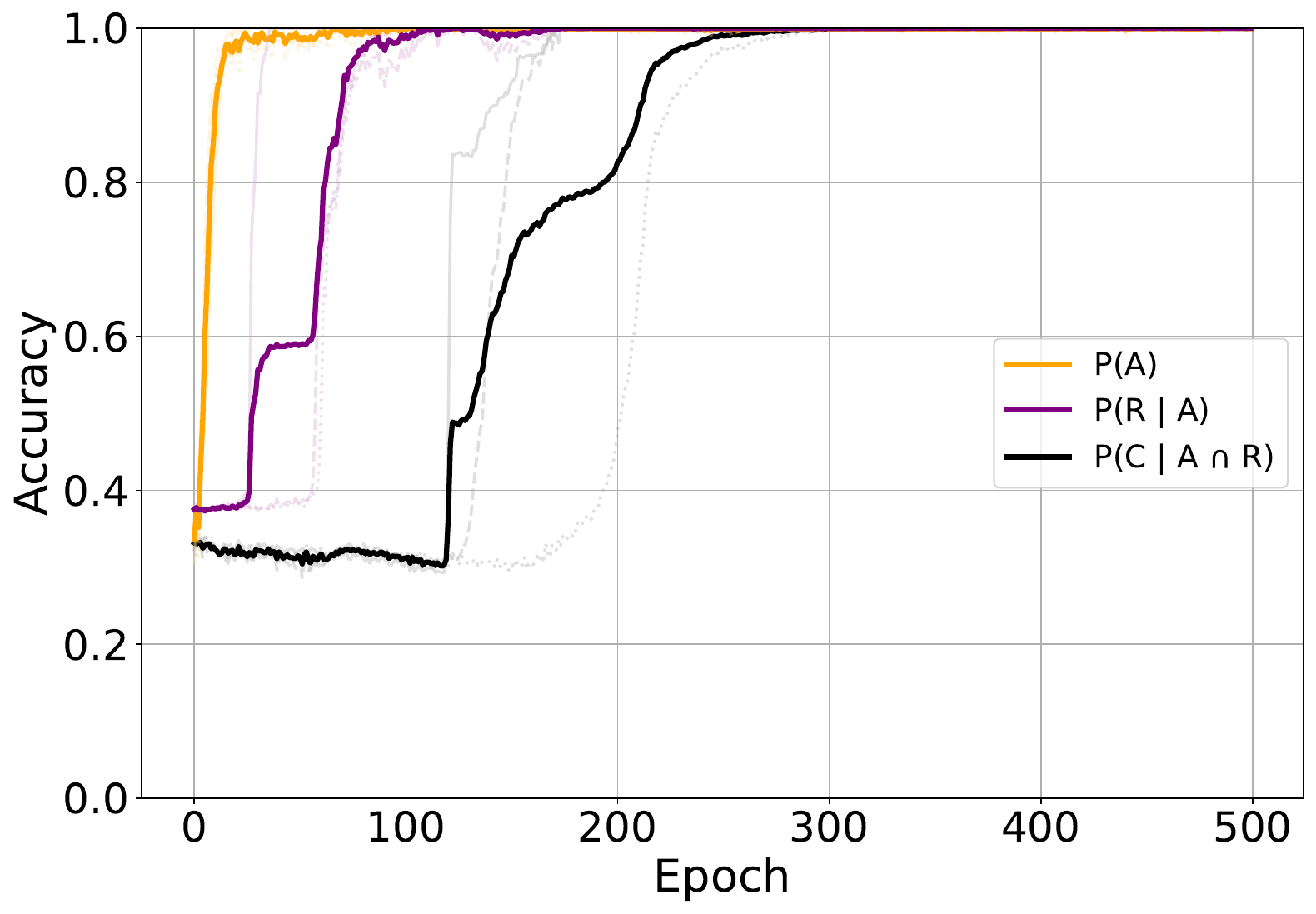}
        \caption{$hl_{query}=4$}
        \label{fig:composition_4_hop_phasic_learning}
    \end{subfigure}
    \hfill
     \begin{subfigure}[t]{0.24\linewidth}
        \centering
        \includegraphics[width=\linewidth]{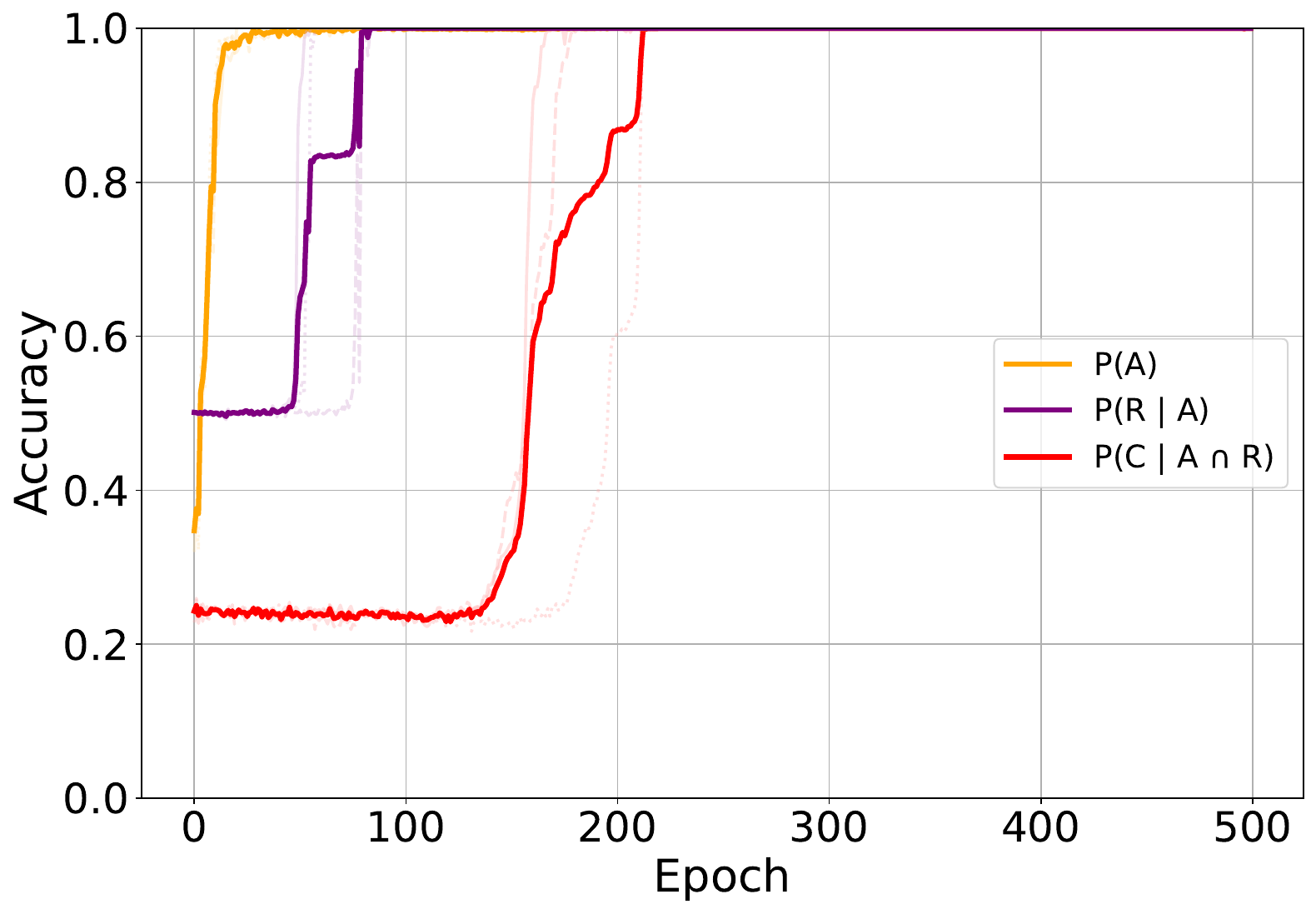}
        \caption{$hl_{query}=5$}
        \label{fig:composition_5_hop_phasic_learning}
    \end{subfigure}
    \caption{Staged learning dynamics for composition. We plot $\mathbb{P}[A]$ (orange, in-support), $\mathbb{P}[R| A]$ (purple, within reachable stones), and $\mathbb{P}[C\mid A\cap R]$ (exact match) shown as blue, green, black, and red, for $hl_{query} = [2,3,4,5]$ respectively. $\mathbb{P}[A]\;\mathbb{P}\;[R| A]\;\mathbb{P}[C| A\cap R]$ recovers the overall accuracy in Figure \ref{fig:composition_performance_all_hops}. Lighter and darker curves denote individual runs and the mean respectively.}
    \label{fig:composition_phasic_learning}
\end{figure*}

To study the stages, we define three meaningful events that naturally follow from the composition task structure. Given a chemistry $chem$, query $q_i$, and hop length $k$, we define: 
\begin{align*}
A &:= \{\hat y_i \in T_{chem}\} &&\text{(in-support)},\\
R &:= \{\hat y_i \in R_k(x_i)\} &&\text{(reachable stones)},\\
C &:= \{\hat y_i = y_i\} &&\text{(exact match)}.
\end{align*}
$A$ and $C$ are as defined previously. $R_k(x_i)$ defines the set of stones ``reachable'' in $k$ hops transitions from the query stone $x_i$. For instance, in Figure \ref{fig:example_of_a_chemistry_graph_composition_decomposition} (left panel), if $x_i = S1$, and $k=hl_{query}=2$, the target $y_i$ can only be one of  $R_2(x_i) =\{S3, S8, S5\}$. Because by construction, $y \in R_k(x_i) \subset T_{chem}$, we have $C \subset R \subset A$, and thus, 
$\mathbb{P}[C] = \mathbb{P}[A]\; \mathbb{P}[R| A]\; \mathbb{P}[C| A\cap R]$. We now describe the stages in Figure \ref{fig:composition_phasic_learning}. Across hop lengths, the model first restricts its prediction to the support set, then learns the reachable set, and finally the exact target.

\begin{enumerate}
    \item \emph{Stage 1: In-support.}
    $\mathbb{P}[A]$ (orange) rises rapidly to perfect accuracy, correctly restricting the predictions to the 8 support stones. This stage mirrors ``in-support'' behavior (Section \ref{sec:analytical_study_4_edge_held_out}).
    
    \item \emph{Stage 2: Within reachable stones given in-support}.
    $\mathbb{P}[R | A]$ rises from chance to $\sim$100\%, as the model identifies the $k$-hop stones reachable from $x_i$. We note that chance levels (purple curves in Figure \ref{fig:composition_phasic_learning}) vary by hop length as the number of reachable stones changes per hop length: $3/8 = 37.5 \%$ for $k \in \{2,4\}$, and  $4/8 = 50\%$ for $k \in \{3,5\}$. Note that the reachable sets coincide for same hop length parity (i.e., the reachable sets for a given query stone are the same for hops 2/4, and 3/5), but the longer sequences still need to be resolved on the cube.
    Stage 2 is unique to composition, requiring reachable stones identification before disambiguating among them.
    
    \item \emph{Stage 3: Exact match given reachable stones. }
    $\mathbb{P}[C| A\cap R]$ reaches $\sim$100\% as the model learns to correctly disambiguate the target among the reachable set given $x_i$ and $z_i$ and $k$. 
\end{enumerate}
\par

This sequential progression reveals a \textit{coarse-to-fine} learning behavior --- acquiring broad constraints before exact targets --- grounding prior findings \citep{arpit2017closer, gopalaniWhatHappensLoss2025} in explicit latent structure.

\section{Delayed Learning in Decomposition}
\label{sec:intermediate_stone_state_inference_experiment}

We now study the difference in model convergence when systematically varying $hl_{support} \in \{2,3,4,5\}$ to solve 1-hop queries ($hl_{query}=1$). Increasing $hl_{support}$ tests the model's ability to infer the underlying chemistry from multi-hop potion sequences to solve a 1-hop query. In our experiments, we provide all the examples in the support for a given $hl_{support}$, and similar to the composition experiments (Section \ref{sec:composition_experiment}), we exclude support trajectories with cycles.

\par
\begin{figure}[!htb]
    \centering
    \includegraphics[width=0.5\linewidth]{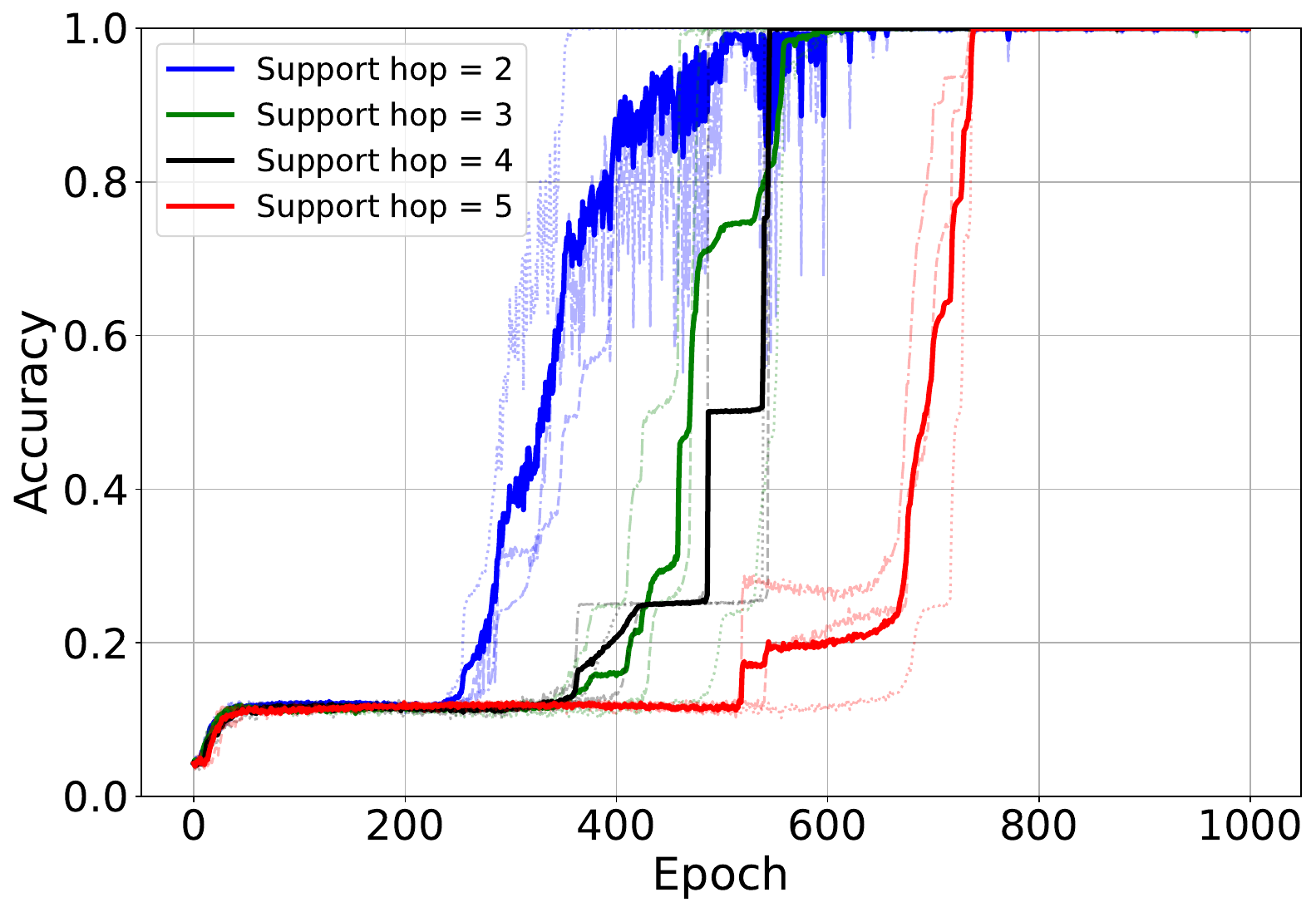} \caption{Overall decomposition performance (1000 epochs). We observe delayed model convergence with increasing $hl_{support}$ with 2-hop converging the earliest and 5-hop converging the latest. The darker curves denote the mean accuracy and the lighter curves show the individual runs.
    }
    \label{fig:decomposition_performance_all_hops}
\end{figure}

We show the decomposition results in Figure \ref{fig:decomposition_performance_all_hops}. We
observe that under fully enumerated support, increasing $hl_{support}$ delays model convergence, with $hl_{support}=2$ reaching convergence sooner than $hl_{support}=5$.
This finding echoes recent works on understanding language model performance under increasing complexity \citep{dziriFaithFateLimits2023, shojaeeIllusionThinkingUnderstanding2025}, and extends it to a setting to study latent structure learning. We also observe multiple stages across $hl_{support} \in \{2,3,4,5\}$, which we investigate next. \par

\subsection{Analysis of decomposition learning stages}
\begin{figure*}[!htb]
    \centering
    \begin{subfigure}[t]{0.24\linewidth}
        \centering
        \includegraphics[width=\linewidth]{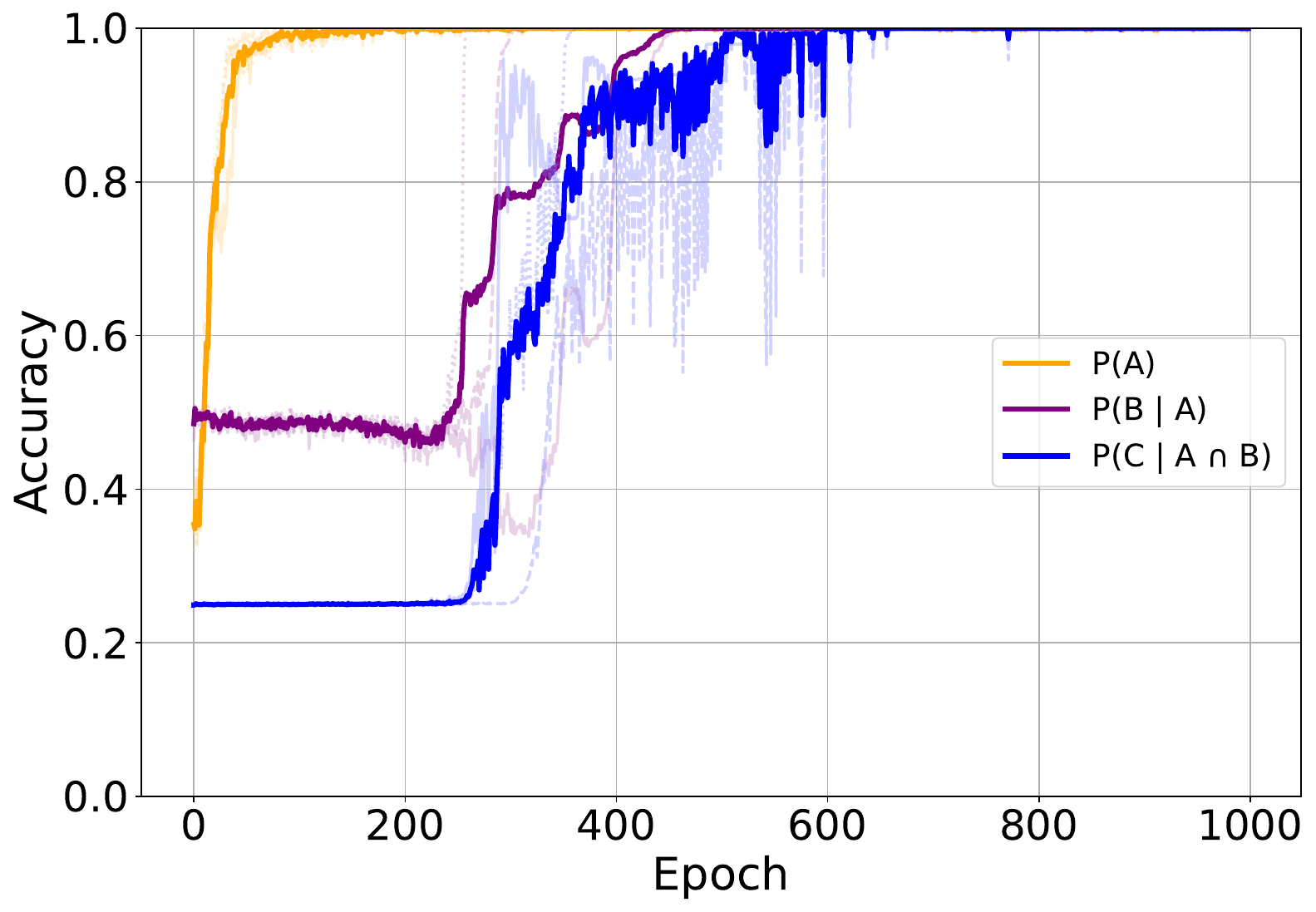}
        \caption{$hl_{support} = 2$}
        \label{fig:decomposition_2_hop_phasic_learning}
    \end{subfigure}
    \hfill
    \begin{subfigure}[t]{0.24\linewidth}
        \centering
        \includegraphics[width=\linewidth]{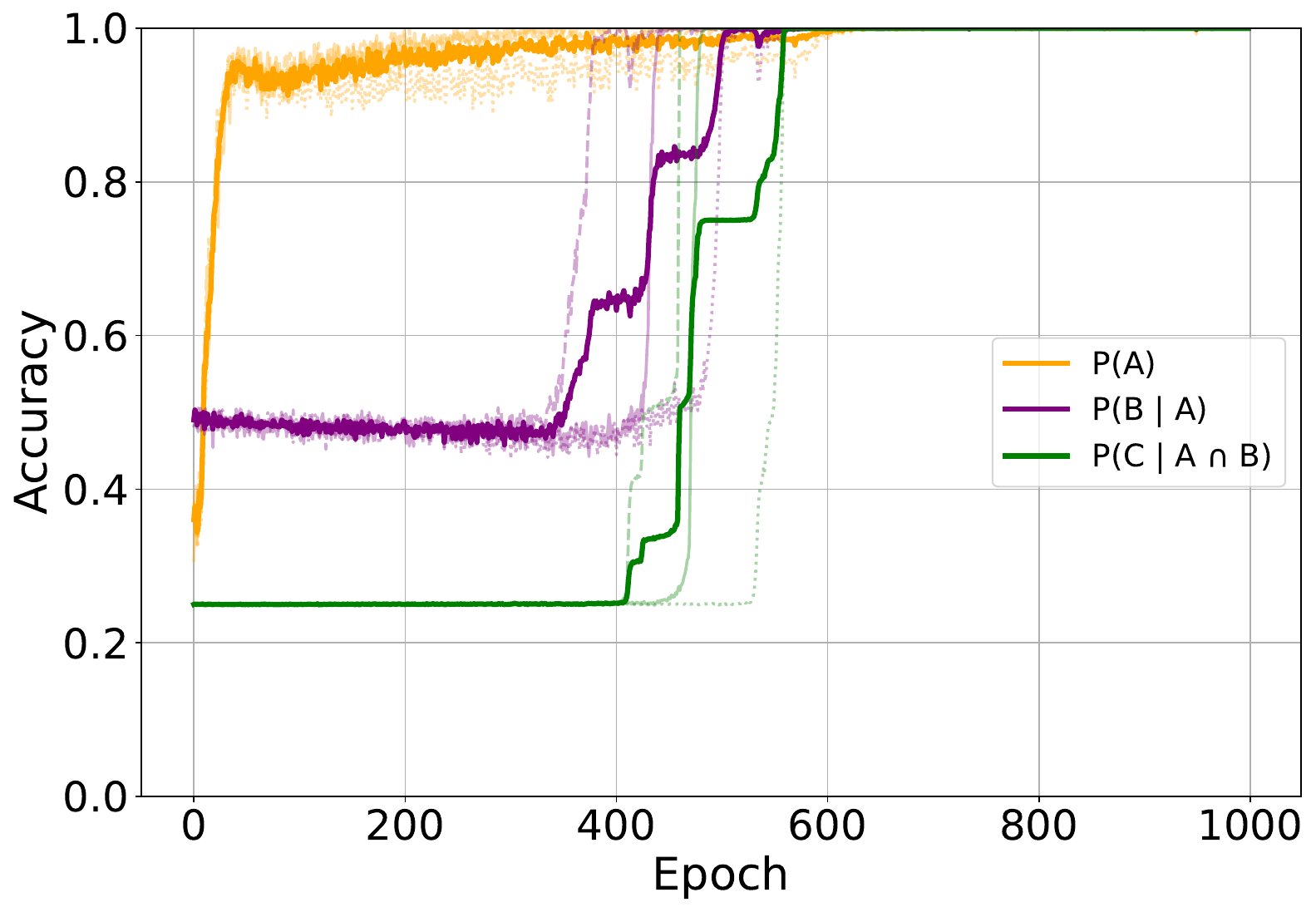}
        \caption{$hl_{support} = 3$}
        \label{fig:decomposition_3_hop_phasic_learning}
    \end{subfigure}
    \hfill
    \begin{subfigure}[t]{0.24\linewidth}
        \centering
        \includegraphics[width=\linewidth]{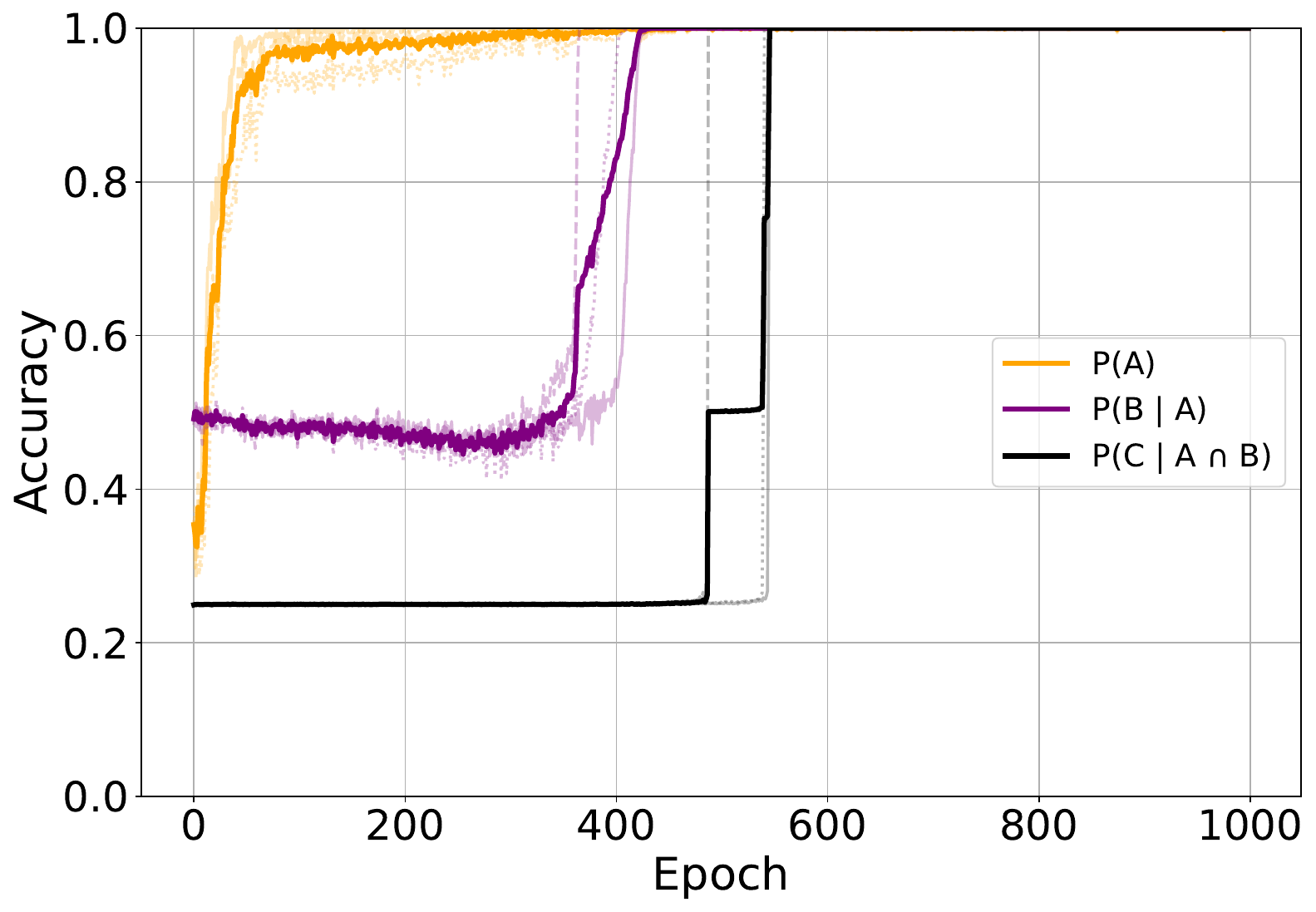}
        \caption{$hl_{support} = 4$}
        \label{fig:decomposition_4_hop_phasic_learning}
    \end{subfigure}
    \hfill
     \begin{subfigure}[t]{0.24\linewidth}
        \centering
        \includegraphics[width=\linewidth]{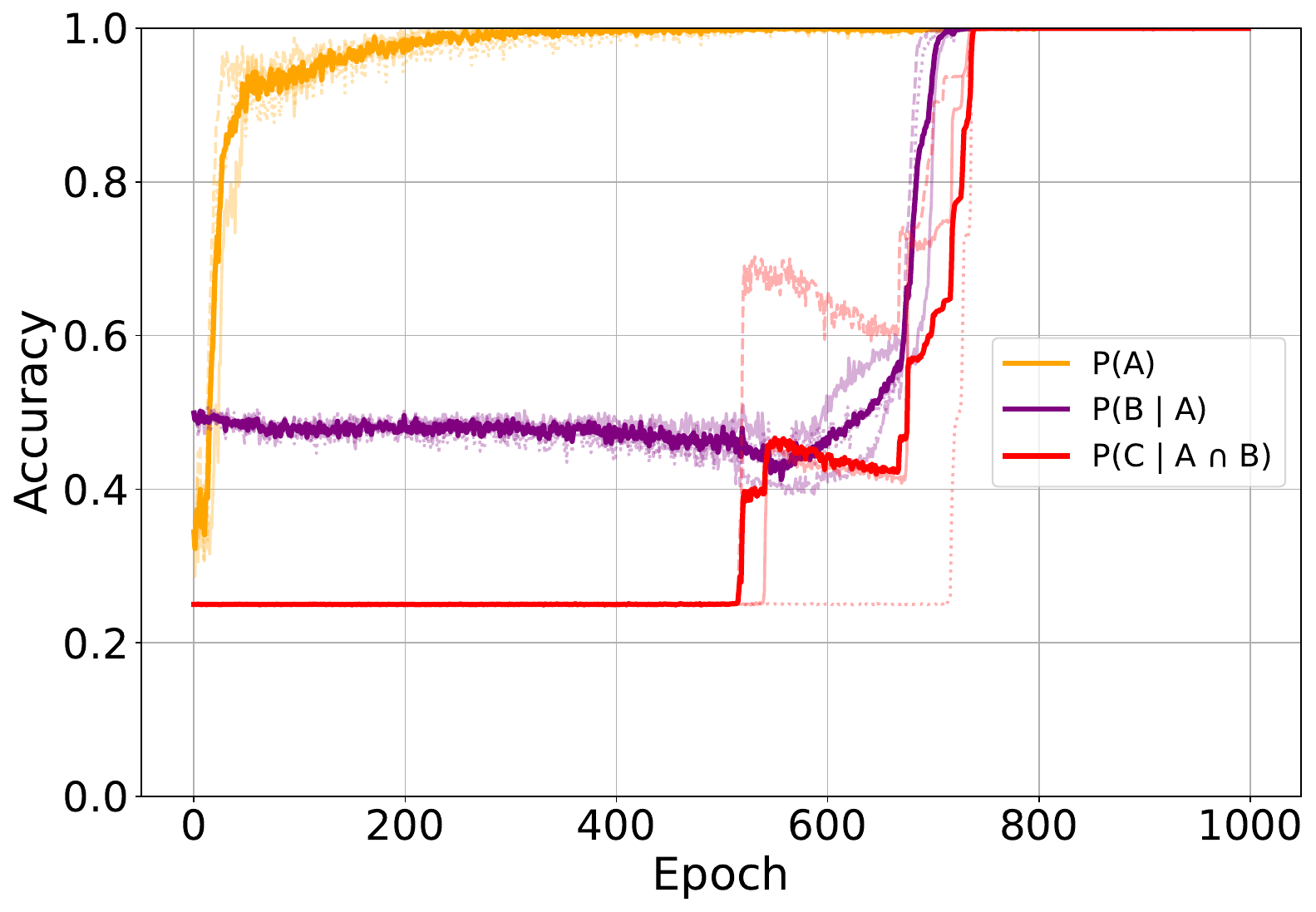}
        \caption{$hl_{support} = 5$}
        \label{fig:decomposition_5_hop_phasic_learning}
    \end{subfigure}
    \caption{Staged learning dynamics for decomposition. Components include $\mathbb{P}[A]$ (orange, in-support), $\mathbb{P} [B \mid A]$ (purple, reachable by $z_i$), and $\mathbb{P}[C| A \cap B]$ (exact match) shown as blue, green, black, red, for $hl_{support} = [2,3,4,5]$ respectively;  $\mathbb{P}[A]\;\mathbb{P}\;[B| A]\;\mathbb{P}[C| A\cap B]$ product recreates the overall accuracy in Figure \ref{fig:decomposition_performance_all_hops}. Lighter and darker curves denote the individual runs and the mean.} 
    \label{fig:decomposition_phasic_learning}
\end{figure*}

For a given query $q_i$ containing the stone $x_i$ and potion $z_i$, the target $y_i$ belongs to the set of four stones reachable by $z_i$. We denote this set as the ``reachable by $z_i$'' because applying $z_i$ can only result in four stones. This is a meaningful event because the final potion determines the set of possible end stone states.
For example, in Figure \ref{fig:example_of_a_chemistry_graph_composition_decomposition} (left), applying a yellow 1-hop potion to $x_i = S1$ can only result in a stone from the set of destination stones $\{S6, S5, S4, S3\}$. In other words, there are four stones where a potion $z_i$ (e.g., yellow) is an incoming transition. We define:
\begin{align*}
A &:= \{\hat y_i \in T_{chem}\} &&\text{(in-support)},\\
B &:= \{\hat y_i \in H_{chem}^{(reachable)}\} &&\text{(reachable by $z_i$)},\\
C &:= \{\hat y_i = y_i\} &&\text{(exact match)}.
\end{align*}
the product of which gives the exact match accuracy
$\mathbb{P}[C] =\mathbb{P}[A]\mathbb{P}[B | A]\mathbb{P}[C|A \cap B]$.
We now describe the stages shown in Figure \ref{fig:decomposition_phasic_learning}.

\begin{enumerate}
    \item \emph{Stage 1: In-support.}
    Across $hl_{support} \in \{2,3,4,5\}$, the model quickly learns to restrict its predictions to the support set, as shown by the rapid rise of $\mathbb{P}[A]$ to near perfect accuracy. 
    
    \item \emph{Stage 2: Reachable by $z_i$ given in-support.} 
    For $hl_{support} \in \{2,3,4,5\}$, $\mathbb{P}[B \mid A]$ remains at chance ($\approx 50\%$) for many epochs, indicating that the model cannot learn this stage immediately after learning $\mathbb{P}[A]$. This is reflected by the stagnation segment at 12.5\% in Figure \ref{fig:decomposition_performance_all_hops}, showing that the model struggles to advance beyond Stage 1, to learn Stage 2.
    
    \item \emph{Stage 3: Exact match given reachable by $z_i$.}
    For $hl_{support} \in \{2,3,4,5\}$, we observe that after learning Stage 2, the exact match accuracy rises quickly, i.e., the model correctly learns the target from the set of four stones. This is reflected in the rapid rise of $\mathbb{P}[C | A \cap B]$, from chance to perfect accuracy.
\end{enumerate}

In summary, increasing $hl_{support}$ primarily delays the acquisition of $\mathbb{P}[B \mid A]$ and subsequent model convergence. Nonetheless, the stages are learned in a coarse-to-fine sequence: $\mathbb{P}[A]$, followed by $\mathbb{P}[B \mid A]$, and finally $\mathbb{P}[C \mid A \cap B]$. To understand why $\mathbb{P}[B \mid A]$ is delayed, we investigated if other intermediate stages exist (e.g., initially narrowing the search space to candidate stones). However, the simultaneous rise of $\mathbb{P}[B \mid A]$ and other relevant metrics indicated their absence (Appendix \ref{appendix:intermediate_stages_decomposition}).
\par

\textbf{Hyperparameter Sensitivity:} While increased complexity delays latent structure learning, we also observed hyperparameter sensitivity that alters staged dynamics or prevents the convergence of later stages within the compute budget. We discuss such failure modes in Appendix \ref{appendix:hyperparameter_sensitivity} and the stage-wise success rates in Table \ref{tab:stage_success_rates_per_task_hop_length}.

\paragraph{Causal Evidence for Stage Ordering:} To provide causal evidence for the coarse-to-fine stage ordering, we conducted layer freezing experiments to test whether later stages depend on continued plasticity of particular layers. Starting at a given epoch, we freeze one layer (one of the four transformer layers, or the token embedding layer) keeping the rest of the model unfrozen. Then, we measure the delay in acquiring a subsequent stage relative to the first stage ($\mathbb{P}[A]$). We identify layer specific plasticity windows, during which freezing a transformer layer delays (and sometimes prevents) later stage acquisition (results in Appendix \ref{sec:plasticity_window}). 




\par


\section{Related Work}
\label{sec:related_works}

\textbf{Latent structure learning:} Recent work has shown that transformers encode graph-like latent structure \citep{nichaniHowTransformersLearn2024, hendersonTransformersGraphtoGraphModels2023}, and that attention heads in models like BERT \citep{devlinBERTPretrainingDeep2019} encode linguistic structures \citep{ manningEmergentLinguisticStructure2020, ravishankarAttentionCanReflect2021}.
However, analyzing the learning progression is difficult due to the complexities of natural language. To this end, we conducted experiments on a controlled benchmark to analyze latent structure learning dynamics of transformer models.

Transformer learning often proceeds in stages \citep{barakHiddenProgressDeep2022, chenTrainingDynamicsMultiHead2024, gopalani2024abrupt, chenSuddenDropsLoss2025, gopalaniWhatHappensLoss2025, songOutofdistributionGeneralizationComposition2025, zhangTrainingDynamicsInContext2025}, and is frequently linked to ``grokking'' \citep{power2022grokkinggeneralizationoverfittingsmall, nandaProgressMeasuresGrokking2023} and the broader ``emergence'' phenomenon \citep{weiEmergentAbilitiesLarge2022}.
Other studies connect loss plateaus and abrupt learning to task features \citep{gopalani2024abrupt, chenSuddenDropsLoss2025}, gradient-based dynamics \citep{saxeMathematicalTheorySemantic2019, nakkiranSGDNeuralNetworks2019, zhangSaddletoSaddleDynamicsExplains2025b}, while mechanistic analyses link abrupt learning to circuit formation \citep{reddyMechanisticBasisData2023, olssonIncontextLearningInduction2022, singhWhatNeedsGo2024}. Prior work has also studied such behavior from the perspective of hierarchical structure \citep{abbeStaircasePropertyHow2021}, hidden progress in learning dynamics \citep{barakHiddenProgressDeep2022}, multi-stage in-context learning dynamics \citep{edelmanEvolutionStatisticalInduction2024}, and composition subskill acquisition order \citep{zhao2026shattered}. \citet{daiUncoveringGraphReasoning2025} investigate graph reasoning via circuit analysis, but do not focus on learning dynamics. Such settings typically do not provide an explicit latent structure that allows stages to be attributed to specific latent structure components. 
We use a controlled setting to bridge this gap and show that accuracy jumps correspond to the sequential acquisition of distinct latent structure components.



\textbf{Composition:} Composition remains a challenge for transformers, especially given increasing task complexity \citep{hupkesCompositionalityDecomposedHow2020, dziriFaithFateLimits2023, pettyRELICEvaluatingCompositional2025} and compositional generalization \citep{ontanonMakingTransformersSolve2022a, kobayashiWhenCanTransformers2024}. While some works train models from scratch on compositional tasks \citep{thommLimitsTransformerLanguage2024}, they focus on final learned models, and tasks without clear latent structures, making learning behavior difficult to analyze. Recent works explore answering multi-hop queries from 1-hop transitions \citep{allen-zhuPhysicsLanguageModels2025, musatMechanismEmergenceStacked2025, wangLearningCompositionalFunctions2025}, typically finding that performance degrades as hop length increases. In contrast, our Alchemy setup reveals no increase in convergence with increasing hop length. We hypothesize that this difference arises from the underlying graph topology: Alchemy stones have three outgoing transitions on a cubic graph, differing from other setups.
\par
\textbf{Decomposition:} Recent decomposition research focuses on LLM prompting strategies \citep{khotdecomposedpromptingmodular2023, pasewarkretuningovercomingcompositionality2024a, wei2023chainofthoughtpromptingelicitsreasoning, prasadadaptasneededdecomposition2024}, leaving its staged dynamics during latent structure learning largely unexplored. In our setup, increasing support hop length delays stage convergence. This mirrors \citet{zucchetemergencesparseattention2025}, who show that data complexity delays phase transitions during the emergence of sparse attention. Mechanistically connecting the stage convergence delay by monitoring the attention patterns remains a promising avenue 
for future work.

\section{Conclusion}
\label{sec:conclusion}
Our results demonstrate how transformers acquire latent structure through controlled experiments using the Alchemy benchmark. By factorizing model accuracy into interpretable events grounded in specific well-defined latent structures, we identified which components are learned at discrete training stages and showed that a transformer model learns in a \textit{coarse-to-fine} manner. The model first narrows down its predictions to broad subsets of valid outcomes before refining them to exact targets. In the withheld-potion task, our analysis also revealed a reward-based adjacency bias that explains the counterintuitive order of the learning stages. Through controlled inference-time ablation studies to investigate the counterintuitive stage ordering, we showed that when the reward  structure is masked or replaced, the model uses an alternative intermediate latent structure to learn the task. Across tasks with varying complexity, we found an asymmetry between composition and decomposition: composition model convergence and stage timing do not noticeably vary with increasing query hop length, whereas decomposition convergence is delayed as support hop length increases, which is reflected by a delayed later-stage acquisition.
Together, these results show how a transformer acquires latent structure through interpretable stages in a controlled setting, providing a detailed understanding of how different capabilities evolve during training. 

\section{Limitations}
\label{appendix:limitations}
While the controlled experiments in this work provide insights into how transformers learn latent structure, our conclusions are based on a small decoder-only transformer (\(\sim\)2M parameters) trained on the Alchemy benchmark with a latent cube structure. Although this setting allows for precise and depth-first analysis of staged dynamics, it limits the extent to which our findings can be generalized to larger models, different architectures, or less structured latent graphs. Nevertheless, our work provides a sense of the type of learning patterns we might search for in large scale models, which will ultimately lead to more generalizable insights in frontier AI systems. Thus, more work is needed to confirm that the \textit{coarse-to-fine} staged learning appears in larger models and more complex settings. Although we do not localize the internal circuits or representations responsible for the observed stages, our analyses provide a basis for future work on the mechanistic interpretability of latent structure learning \citep{meng2022locating, zhang2024towards, yang2025emergent}. Our work also motivates the development of more benchmarks to study transformer learning dynamics and latent structure acquisition more broadly. 
Future work should also investigate the interplay of optimization algorithms and staged learning of latent structure.
Lastly, developing strategies to mitigate the decomposition convergence delay is an important direction for future work.

\subsubsection*{Broader Impact Statement}
Our work contributes to a better understanding of how a small decoder-only transformer model learns latent structure by analyzing the model's learning dynamics. Our results advance the field of deep learning by improving our understanding of how the capabilities of transformer models emerge during training. Because this work is conducted in a controlled setting with small models and does not introduce any high-risk capability, we believe it is unlikely to have negative societal impacts on its own.



\bibliography{main}
\bibliographystyle{tmlr}

\appendix

\section{Training details}
\label{appendix:training_details_hyperparameter}

We implemented all models using PyTorch \citep{paszke2019pytorchimperativestylehighperformance} version 2.6. For hyperparameter tuning, we conducted a large sweep using Weights \& Biases \citep{wandb} over the configurations detailed in Table \ref{tab:latent_structure_learning_hyperparams}. Each experiment used the AdamW optimizer \citep{loshchilov2018decoupled} and was run on a compute cluster utilizing 4 NVIDIA L40S GPUs (48 GB VRAM) with FP32 precision.

\begin{table}[!htb]
    \centering
    \begin{tabular}{p{3cm}p{3cm}}
        \toprule
        Hyperparameter & Values \\ 
        \midrule
        learning rate & [1e-3, 4e-4, 5e-4, 1e-4, 9e-5, 7e-5, 1e-5]\\ \hline
        scheduler & cosine, cosine w/ restarts\\ \hline
        weight decay & [0.1, 0.01, 0.001]\\ 
        \midrule
        minimum learning rate (cosine and cosine w/ restarts) & 7e-5, 8e-5, 9e-5, 8.5e-5, 9.5e-5, 1e-5\\ \hline
    \end{tabular}
    \caption{Set of hyperparameters swept over for the experiments.}
    \label{tab:latent_structure_learning_hyperparams}
\end{table}

Following an initial evaluation, we fixed the learning rate to 1e-4, as the model failed to reach the first stage (in-support) with other values regardless of the configuration. The choice of scheduler had minimal effect, though the cosine scheduler performed best (we used the cosine scheduler throughout). We subsequently executed each experiment across 3 random seeds, yielding 54 runs per task and hop-length combination (3 seeds $\times$ 3 weight decays $\times$ 6 minimum learning rates). A comprehensive study of how various hyperparameter choices affect stage completion is an important direction for future work.

\begin{table}[h]
\centering
\small
\begin{tabular}{llcc}
\toprule
Task & Hop length & \# support examples & Input length \\
\midrule
Withheld potion pair & $hl_{\text{support}}=1,\ hl_{\text{query}}=1$ & 16 & 181 \\
\midrule
Composition & $hl_{\text{support}}=1,\ hl_{\text{query}}=2$ & 24 & 270 \\
Composition & $hl_{\text{support}}=1,\ hl_{\text{query}}=3$ & 24 & 271 \\
Composition & $hl_{\text{support}}=1,\ hl_{\text{query}}=4$ & 24 & 272 \\
Composition & $hl_{\text{support}}=1,\ hl_{\text{query}}=5$ & 24 & 273 \\
\midrule
Decomposition & $hl_{\text{support}}=2,\ hl_{\text{query}}=1$ & 48  & 581  \\
Decomposition & $hl_{\text{support}}=3,\ hl_{\text{query}}=1$ & 96  & 1253 \\
Decomposition & $hl_{\text{support}}=4,\ hl_{\text{query}}=1$ & 144 & 2021 \\
Decomposition & $hl_{\text{support}}=5,\ hl_{\text{query}}=1$ & 240 & 3605 \\
\bottomrule
\end{tabular}
\caption{Support set sizes and input lengths (in tokens) for each task and hop setting.}
\label{tab:appendix_data_input_lengths}
\end{table}

\paragraph{Compute costs for staged learning experiments:} Training times for individual runs varied based on the task and hop length.
\begin{itemize}
    \item \textbf{Withheld Potion Pair:} $\sim$35 minutes ($\sim$0.58 hours) per run. Total execution time: 54 runs $\times$ 0.58 hours/run $\approx$ 31 hours.
    \item \textbf{Composition ($hl_{query} \in \{2,3,4,5\}$):} $\sim$3 hours per run across 4 distinct hop lengths. Total execution time: 54 runs $\times$ 4 hops $\times$ 3 hours/run = 648 hours.
    \item \textbf{Decomposition ($hl_{support} \in \{2,3,4,5\}$):} Training times varied dynamically by hop length. Total execution time per hop length: 2-hop (54 runs $\times$ 2.75 hours/run $\approx$ 149 hours); 3-hop (54 runs $\times$ 8.5 hours/run = 459 hours); 4-hop (54 runs $\times$ 16.0 hours/run = 864 hours); 5-hop (54 runs $\times$ 42.0 hours/run = 2,268 hours).
\end{itemize}

\textbf{Compute costs for plasticity windows (layer-freezing):} To estimate the computational cost of the layer-freezing experiments, we report an upper-bound calculation using the full 1000-epoch execution times. We optimized the actual run times by continuing the training process from the unfrozen checkpoints. For each task, we froze 5 independent layers across 3 random seeds, stepping in 10-epoch increments between the earliest and latest intervention epochs ( Appendix \ref{sec:plasticity_window}). We report the approximate intervention epoch bounds and the total execution time (upper bound) below:

\begin{itemize}
    \item \textit{Withheld Potion Pair:} Epochs $\sim$ 10 to 200 (20 configurations). Total execution time: 300 runs (20 configs $\times$ 5 layers $\times$ 3 seeds) $\times$ $\sim$0.58 hours/run $\approx$ 174 hours.
    \item \textit{Composition ($hl_{query} \in \{2,3,4,5\}$):} Epochs $\sim$12 to 262 (26 configurations per hop). Total execution time: 1,560 runs (26 configs $\times$ 4 hops $\times$ 5 layers $\times$ 3 seeds) $\times$ 3.0 hours/run = 4,680 hours.
    \item \textit{Decomposition ($hl_{support} \in \{2,3,4,5\}$):} Intervention windows varied dynamically by hop length. Total execution time per hop length:
    \begin{itemize}
        \item \textbf{2-hop:} Epochs $\sim$32 to 382. 540 runs (36 configs $\times$ 5 layers $\times$ 3 seeds) $\times$ 2.75 hours/run = 1,485 hours.
        \item \textbf{3-hop:} Epochs $\sim$41 to 471. 660 runs (44 configs $\times$ 5 layers $\times$ 3 seeds) $\times$ 8.5 hours/run = 5,610 hours.
        \item \textbf{4-hop:} Epochs $\sim$80 to 540. 705 runs (47 configs $\times$ 5 layers $\times$ 3 seeds) $\times$ 16.0 hours/run = 11,280 hours.
        \item \textbf{5-hop:} Epochs $\sim$124 to 744. 945 runs (63 configs $\times$ 5 layers $\times$ 3 seeds) $\times$ 42.0 hours/run = 39,690 hours.
    \end{itemize}
\end{itemize}


\section{Reward-Based Analysis for Withheld Potion Pair}
\label{appendix:adjacency_and_non_support_analysis}

To investigate the effect of adjacency bias (Section \ref{sec:latent_structure_learning}) on the learning curve of $\mathbb{P}[C \mid A \cap B]$, we plot the same events for each individual query stone grouped by reward feature value, $r \in \{+15, +1, -1, -3\}$. 

Alchemy has a specific reward structure such that there is exactly one stone state with reward $+15$ and $-3$ per chemistry, and they lie on the diagonal opposites of the cube (see Figure~\ref{fig:example_of_a_chemistry_graph_composition_decomposition}, left panel). Furthermore, all three stones adjacent to $+15$ have reward $+1$, and all three stones adjacent to $-3$ have reward $-1$.  This defines the reward structure for all 8 vertices of the cube.

\begin{figure*}[!htb]
    \centering
    \begin{subfigure}[t]{0.49\linewidth}
        \centering
        \includegraphics[width=1.0\linewidth]{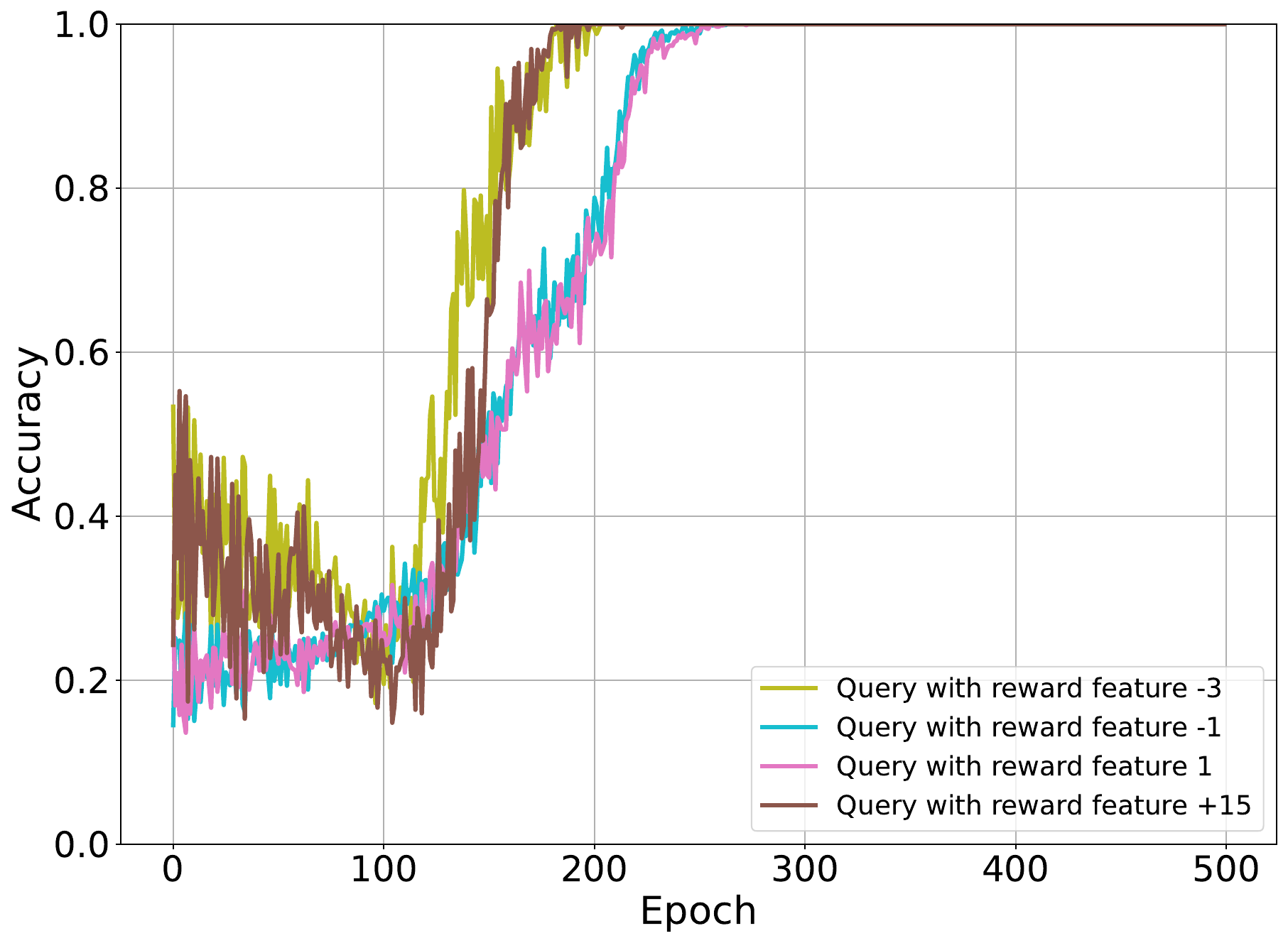}
        \caption{}
        \label{fig:fig:4_edge_held_out_phasic_learning_a}
    \end{subfigure}
    \hfill
    \begin{subfigure}[t]{0.49\linewidth}
        \centering
        \includegraphics[width=1.0\linewidth]{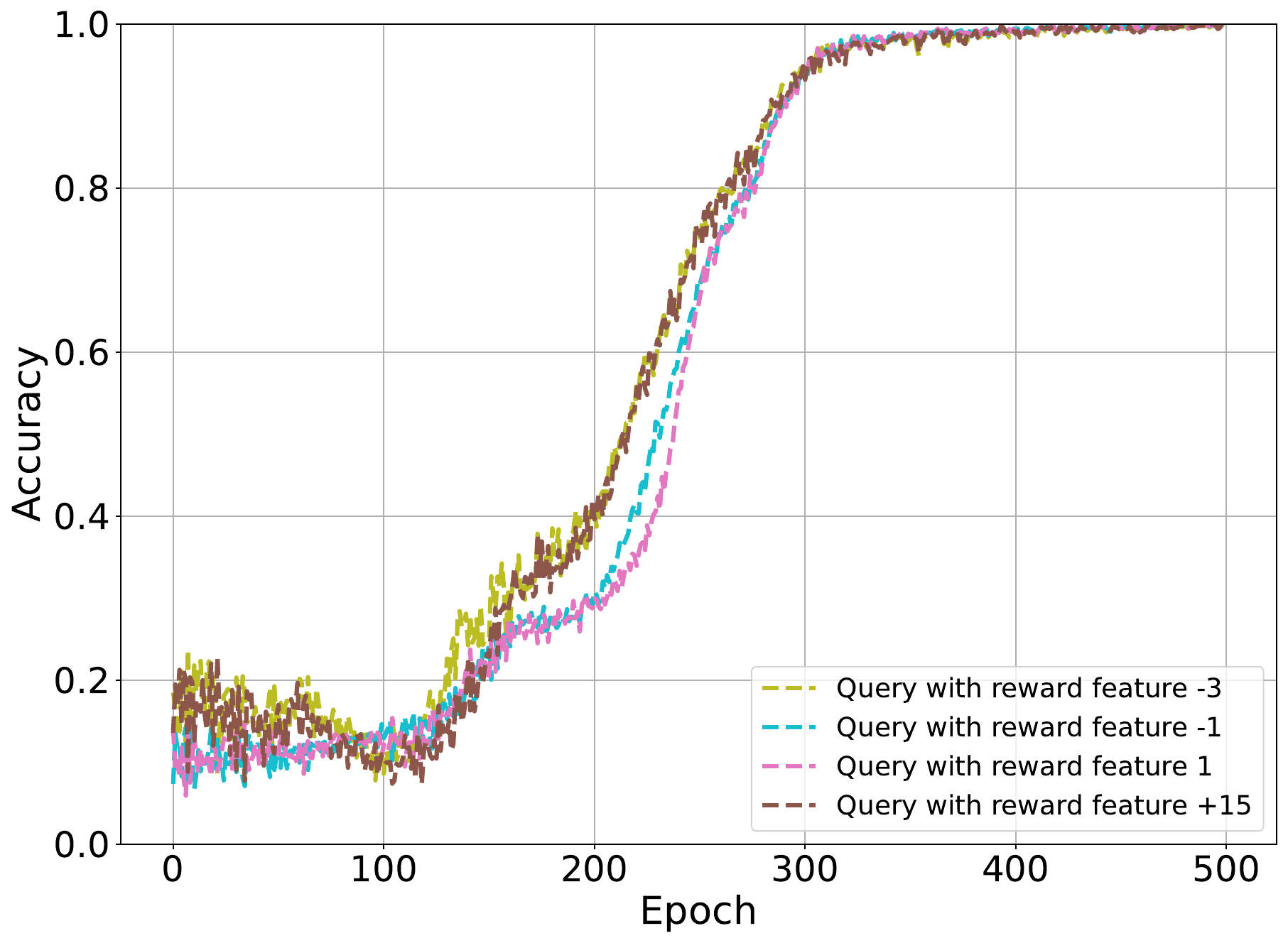}
        \caption{}
        \label{fig:fig:4_edge_held_out_phasic_learning_b}
    \end{subfigure}
    \caption{
    Mean prediction accuracy grouped by the reward value of the query over all stones. (a) Exact match accuracy given correct half. (b) Exact match accuracy given in-support.
    We observe that the model learns the exact match accuracy for queries with -3, and +15 reward features faster than those with -1, and +1 reward features.
    }

    \label{fig:4_edge_held_out_phasic_learning_reward_analysis}
\end{figure*}

From the definition of the events $A$, $B$, and $C$, we can infer that $C \subset B \subset A$, and thus $\mathbb{P}[C \mid A \cap B]$ can be equivalently written as $\mathbb{P}[C \mid B]$. For a query stone, 
Figure \ref{fig:fig:4_edge_held_out_phasic_learning_a} shows the mean accuracy for $\mathbb{P}[C \mid B]$ over all chemistries, grouped by the query stone reward value: $r=+15$ (brown), $r=-3$ (yellow), $r=+1$ (pink), and $r=-1$ (blue). We observe that for $r \in \{-3, +15\}$, $\mathbb{P}[C \mid B]$ rises faster than $\mathbb{P}[C \mid B]$ for $r \in \{-1, +1\}$.
Similarly, Figure \ref{fig:fig:4_edge_held_out_phasic_learning_b} shows the correct target prediction for the in-support stones, $\mathbb{P}[C \mid A]$, for query start stones $x_r$ with distinct reward values. The learning curves again show faster convergence for $r \in \{-3, +15\}$ compared to $r \in \{-1, +1\}$. Through the remainder of this section, we explain these observations further by analyzing the reward-based adjacency bias that exists in the underlying Alchemy latent structure.

\subsection{Staged Learning of Withheld Potion Pair Based on Reward-Adjacency}
\label{appendix:reward_adjacency_grouped_by_reward}

According to Alchemy's underlying latent structure, the reward feature can change by one level per potion transition. For example, from stones with reward $r \in \{+15, -3\}$, transitions are only possible to stones with rewards $+1$ and $-1$, respectively. As a result, stones with rewards $+1$ and $-1$ are adjacent to stones with higher or lower rewards: $\{+15, -1\}$ and $\{+1, -3\}$, respectively.
\par
Consider a query $q = (x, z)$, where $x$ denotes the query start stone, $z$ the query potion, with the target $y$. 
Let $\mathcal{T}_{r}$ denote the set of stones in the graph whose reward feature values are valid for a 1-hop transition from $x_r$. Under the reward constraints above, $\mathcal{T}_{+15}$ and $\mathcal{T}_{-3}$ each contain only three stones with rewards $+1$ and $-1$, respectively. 
However, for $x_r$ with $r \in \{-1, +1\}$, ambiguity arises: a stone with reward +1 may be adjacent to either +15 or -1, resulting in $|\mathcal{T}_{+1} = 4|$. This ambiguity also exists for  a stone with reward -1, resulting in $|\mathcal{T}_{-1}| = 4$.
\par
\begin{figure*}[!htb]
    \centering
    \includegraphics[width=\linewidth]{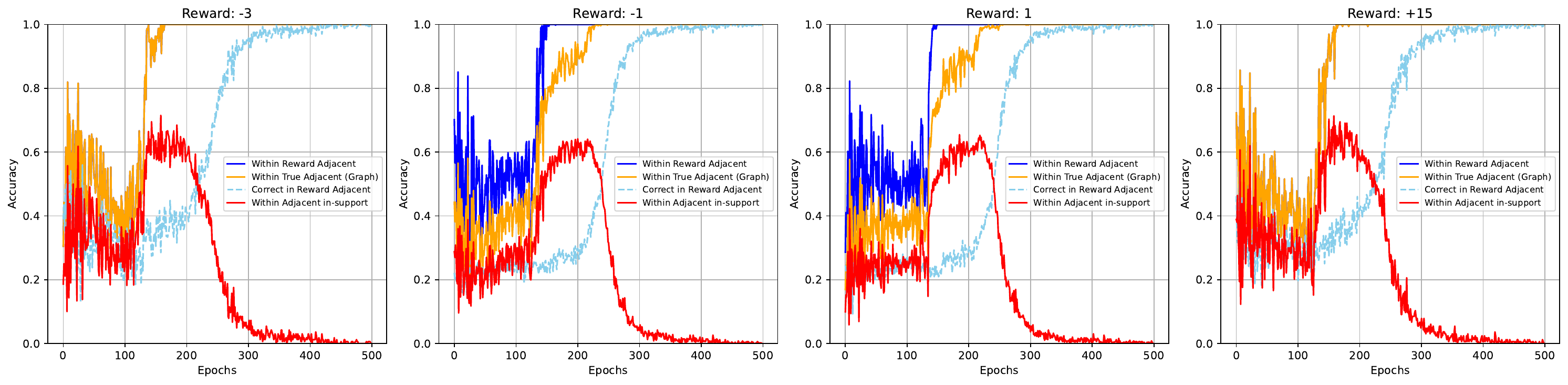}
    \caption{Adjacency analysis for each individual reward feature value: i) dark blue shows the learning curve of the adjacent candidate vertices based on only the reward-related latent structure of the 1-hop transitions (correct set of $\mathcal{T}_r$); ii) orange demonstrates the geometrical adjacency ($\mathcal{N}_r$) subset of neighboring vertices, which overlaps the reward adjacency subset, when the start state has reward values of $\{-3, +15\}$; iii) light blue belongs to the learning of the correct target prediction given the reward-based adjacent subset (correct given $\mathcal{T}_r$); iv) red depicts the selection probability of in-support geometrical 1-hop vertices $\mathcal{R}_r$ (only two vertices in the withheld potion task), which are in the wrong half and should not get selected (goes to zero) as the model fully learns the task. 
    }
    \label{fig:4_edge_held_out_reward_binning_analysis_adjacency}
\end{figure*}

If the model exploits this reward-based structure as a distinguishing signal for predicting the correct half given the query start stone $x_r$, we should see an almost uniform distribution over predictions $\hat y \in \mathcal{T}_{r}$. We test this by plotting $\mathbb{P}_{r}[\hat y = y \mid \hat y \in \mathcal{T}_{r}]$ for $x_r$ with $r \in \{-3, -1, +1, +15\}$. We show the results in Figure \ref{fig:4_edge_held_out_reward_binning_analysis_adjacency} (light blue - dashed), where we observe that during the initial learning period, the values are approximately $\frac{1}{3}$ for $r \in \{-3, +15\}$, and $\frac{1}{4}$ for $r \in \{-1, +1\}$.
\par
We also track how the model learns to identify $\mathcal{T}_{r}$ (within reward adjacent) as the reward-based prediction candidate set within the in-support stones $S$. We measure this by $\mathbb{P}_r[\hat y_r \in \mathcal{T}_{r} \mid \hat y_r \in S]$ and show the results in Figure \ref{fig:4_edge_held_out_reward_binning_analysis_adjacency} (dark blue). During the initial learning period, for $r \in \{-1, +1\}$, $\frac{|\mathcal{T}_{r}|}{|S|} = 4/8 = 0.5$, and for $r \in \{-3, +15\}$, $\frac{|\mathcal{T}_{r}|}{|S|} = 3/8=0.375$. Note that for stones with $r \in \{-3, +15\}$, the model is initially heavily biased towards predicting $\mathcal{T}_{r}$, possibly because of the presence of those transitions in the support $S$.
\par
While $\mathcal{T}_r$ captures the reward-based adjacency, the cubic chemistry structure of Alchemy also exhibits geometric adjacency, where each stone has exactly three geometrically neighboring vertices. We denote this by $\mathcal{N}_r$, and plot the corresponding values (within true adjacent, orange) in Figure \ref{fig:4_edge_held_out_reward_binning_analysis_adjacency}. It naturally follows that for $r \in \{-3, +15\}$, $\mathcal{N}_r$ and $\mathcal{T}_{r}$ are perfectly superimposed because the geometrically neighboring vertices are exactly the reward adjacent vertices.
\par
We further examine the subset of in-support, non-target reward-adjacent stones denoted as $\mathcal{R}_r \subset \mathcal{T}_r$. $\mathcal{R}_r$ contains the reward-adjacent stones that appear in the support $S$ and are in the same half as that of $x_r$, but are not the correct target. Note that for $r \in \{-3, -1, +1, +15\}$ we have $\mid \mathcal{R}_{r} \mid = 2$.
The red curve in Figure \ref{fig:4_edge_held_out_reward_binning_analysis_adjacency} shows this metric as $\mathbb{P}_r[\hat y_r \in \mathcal{R}_{r} \mid \hat y_r \in S]$, where the value rises during training despite these stones being incorrect targets. \par
This mechanism explains the dip in the correct half prediction accuracy observed in Figure \ref{fig:4_edge_held_out_stages} (orange) for the withheld potion effect task (Section \ref{sec:latent_structure_learning}).
The model is initially biased towards predicting the adjacent stones in the same half (until $\approx$ epoch 200) according to the reward adjacency structure. This is reflected in the rise of the red curve in Figure \ref{fig:4_edge_held_out_stages} (predicting the same half as the query stone). As training progresses, the model unlearns to predict any reward-based adjacent stones randomly, and correctly starts predicting the target, which is reflected by the rise in the correct half accuracy.

\section{Investigating the Presence of Intermediate Stages in Decomposition}
\label{appendix:intermediate_stages_decomposition}

In the decomposition experiment (Section \ref{sec:intermediate_stone_state_inference_experiment}), we observed that increasing $hl_{support}$ delayed the learning of $\mathbb{P}[B| A]$ stage. To further investigate this effect, we examined the presence of other intermediate stages.

For the chemistry in Figure \ref{fig:example_of_a_chemistry_graph_composition_decomposition} (left panel), we define $\mathcal{N}(x_i)$ as the set of immediate 1-hop neighbors of the query start stone $x_i$. We then define the following event, for a chemistry $chem$:
\begin{align*}
EN &:= \{\hat y_i \in H_{chem}^{(reachable)} \cup \mathcal{N}(x_i)\} \  \text{(extended neighborhood)}\\
\end{align*}

\begin{figure*}[!htb]
    \centering
    \begin{subfigure}[t]{0.24\linewidth}
        \centering
        \includegraphics[width=\linewidth]{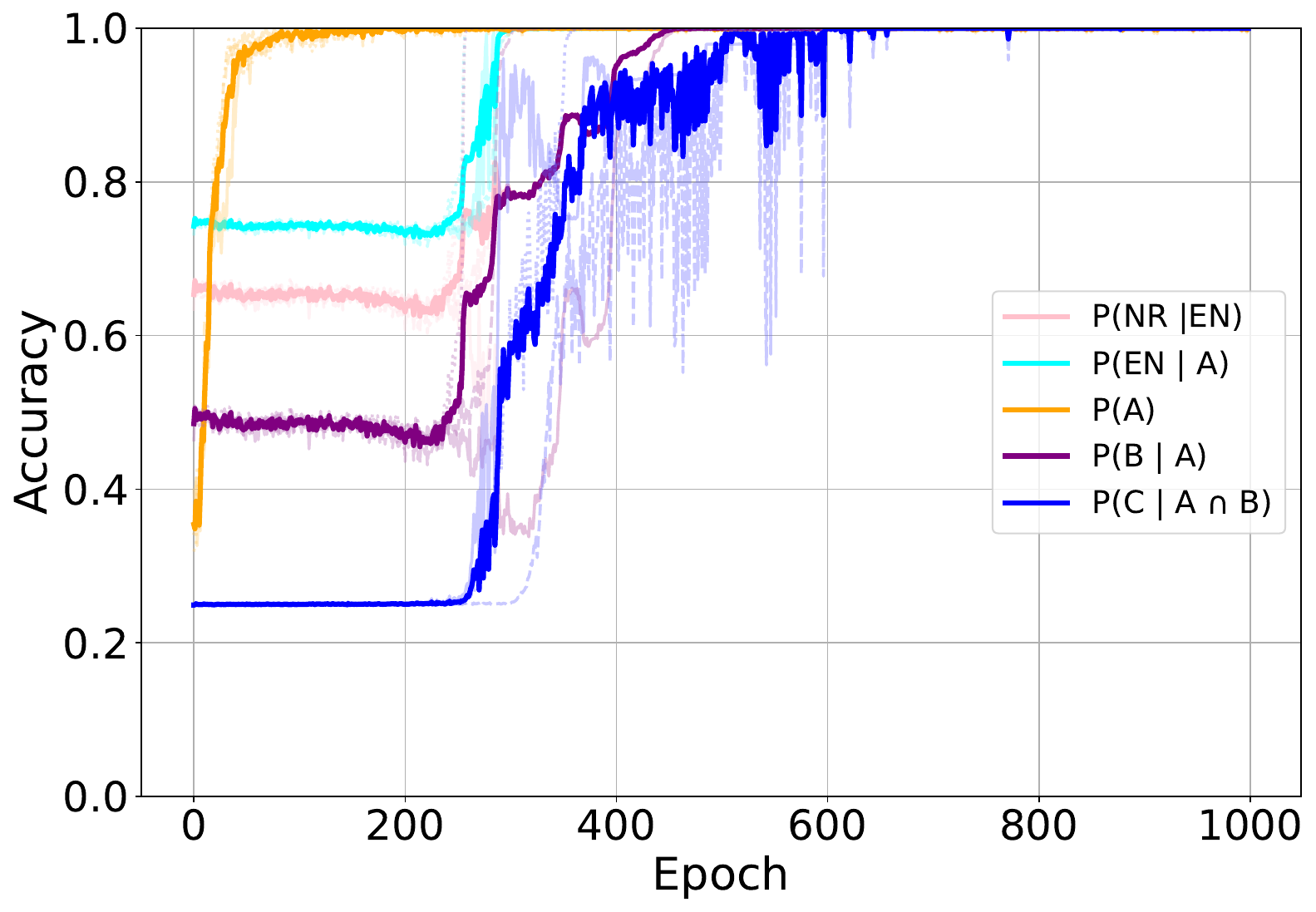}
        \caption{}
        \label{fig:decomposition_with_adjacency_stages_2_hop}
    \end{subfigure}
    \hfill
    \begin{subfigure}[t]{0.24\linewidth}
        \centering
        \includegraphics[width=\linewidth]{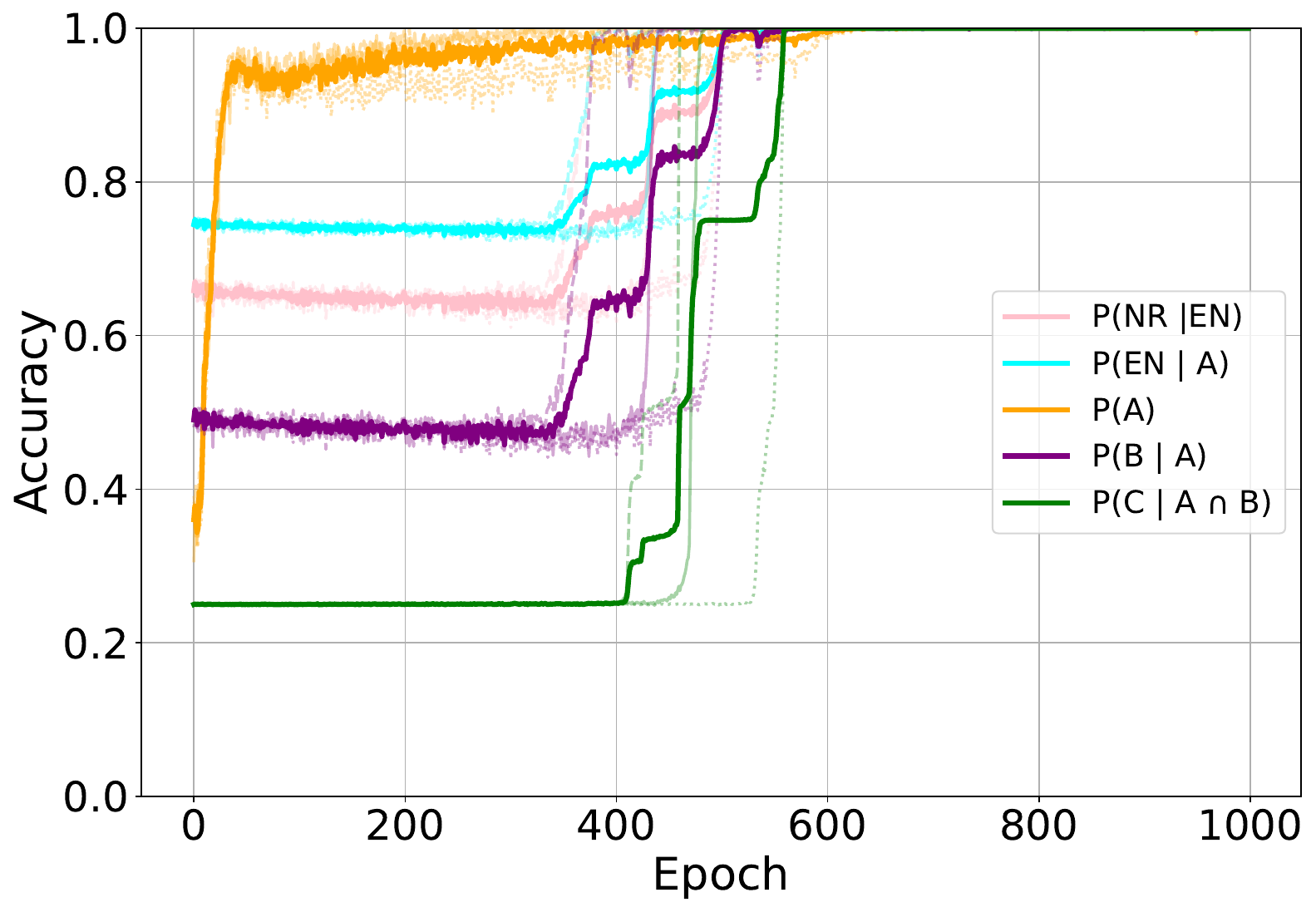}
        \caption{}
        \label{fig:decomposition_with_adjacency_stages_3_hop}
    \end{subfigure}
    \hfill
    \begin{subfigure}[t]{0.24\linewidth}
        \centering
        \includegraphics[width=\linewidth]{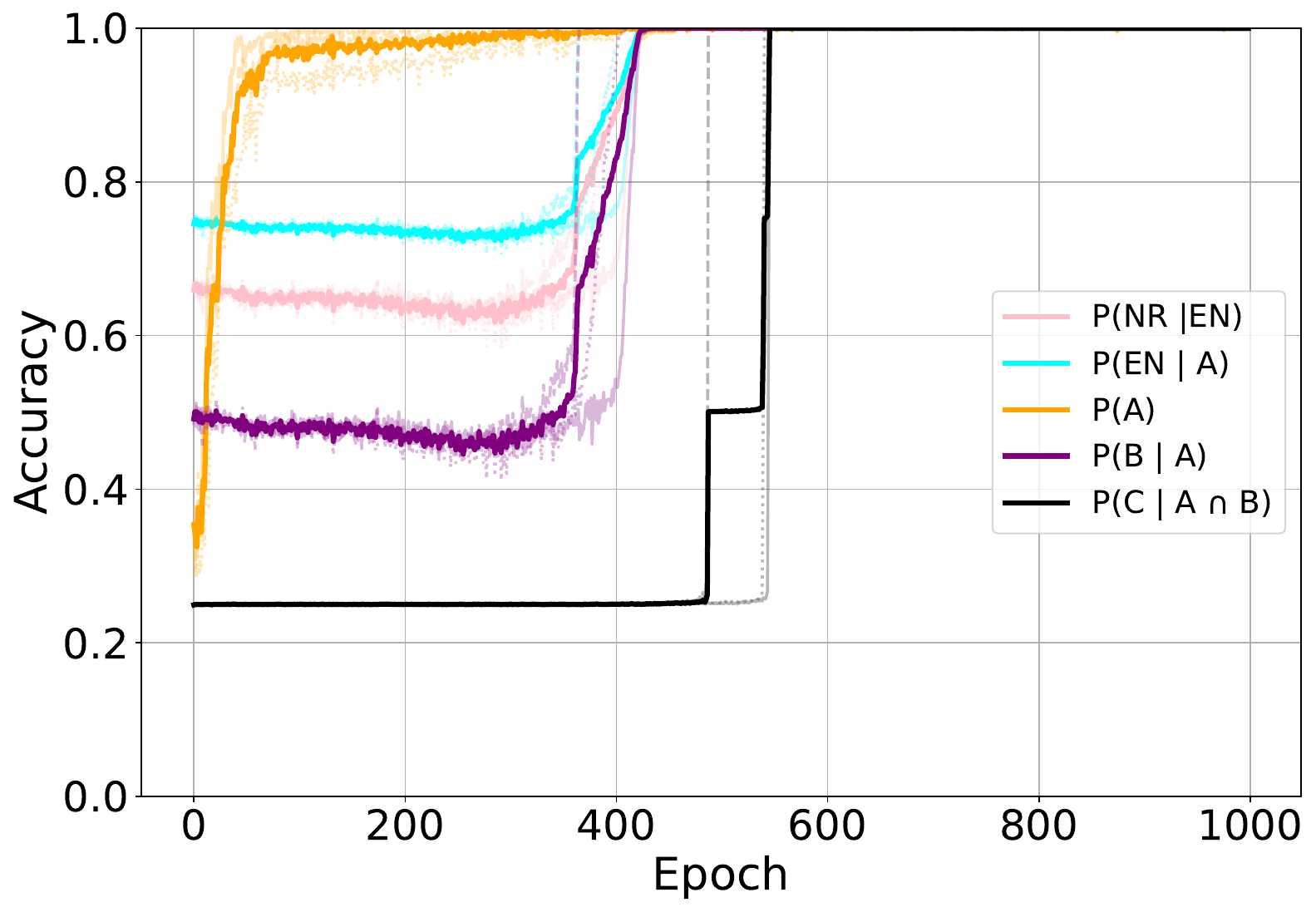}
        \caption{}
        \label{fig:decomposition_with_adjacency_stages_4_hop}
    \end{subfigure}
    \hfill
     \begin{subfigure}[t]{0.24\linewidth}
        \centering
        \includegraphics[width=\linewidth]{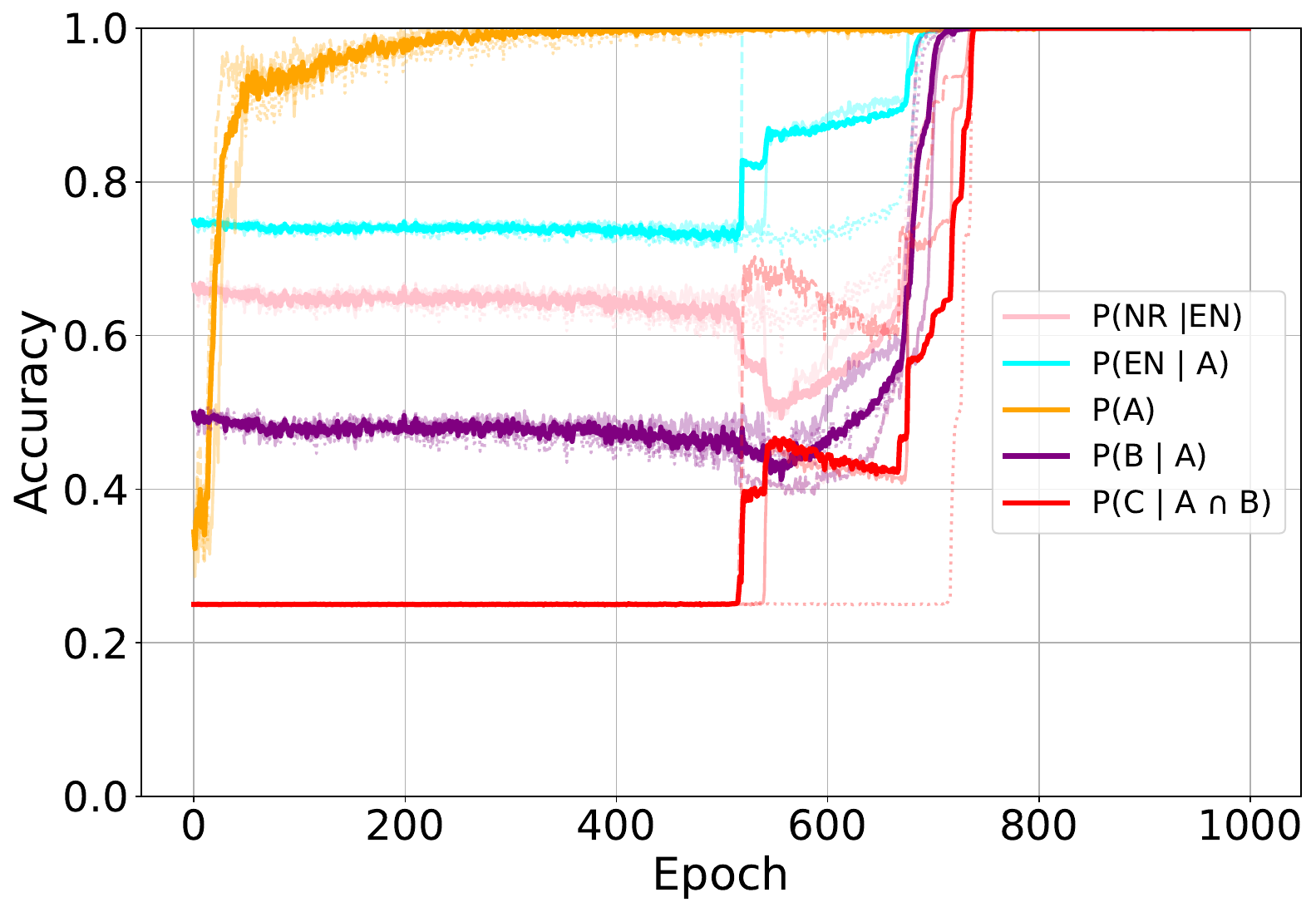}
        \caption{}
        \label{fig:decomposition_with_adjacency_stages_5_hop}
    \end{subfigure}
    \caption{Decomposition staged learning dynamics with the extended neighborhood $EN$ curves and neighborhood refinement $NR$ events. We observed that $EN$ and $NR$ rise simultaneously with the reachable by $z_i$ prediction accuracy $\mathbb{P}[B\mid A]$, suggesting the absence of other intermediate stages.}
    
    \label{fig:decomposition_with_adjacency_phasic_learning}
\end{figure*}

For a given chemistry $chem$, query $q_i$, start stone $x_i$, and the potion $z_i$, event $EN$ (cyan) shows whether the model has narrowed down its predictions to the search space containing the set of stones reachable from $z_i$, and the stones that are the immediate 1-hop neighbors of $x_i$. Note that for a chemistry $chem$, $|H_{chem}^{(reachable)} \cup \mathcal{N}(x_i)| = 6$. The chance accuracy for $\mathbb{P}[EN \mid A]$ is $6/8 = 75\%$.
We also plot the ability of the model to narrow down the search space to reachable stones $H_{chem}^{(reachable)}$ conditioned on learning the extended neighborhood for a given $x_i$. We denote this event by $NR$ (neighborhood refinement, pink). The chance accuracy for $\mathbb{P}[NR \mid EN]$ is $4/6 = 66.67\%$. We show the results in Figure \ref{fig:decomposition_with_adjacency_phasic_learning} for all $hl_{support} = [2,3,4,5]$.

\par
We do not observe any noticeable lag between these events, because the cyan and pink curves rise in immediate succession, indicating a sharp phase transition towards predicting the stones reachable by $z_i$. This indicates that the multi-hop nature of the support $S$ delays the model from successfully extracting the structure of the chemistry, shown by the chance performance of 12.5\% in Figure \ref{fig:decomposition_performance_all_hops} for an extended period of time. This implies that as soon as the model learns that the correct target is in the extended neighborhood, it can almost immediately refine its predictions to the stones reachable by $z_i$.

\section{Case Study on Failure Modes and Hyperparameter Sensitivity}
\label{appendix:hyperparameter_sensitivity}
In our experiments, we observed that the model is sensitive to hyperparameters. We discuss in this section how certain hyperparameters can delay or prevent stage completion. 

\paragraph{Failure modes:} We first provide an overview of the stage success rates for each task / hop length combination. Based on the total number of hyperparameter combinations tested (see Appendix \ref{appendix:training_details_hyperparameter}), we show the success rates for each stage in Table \ref{tab:stage_success_rates_per_task_hop_length}.
\begin{table}[!htb]
\centering
\begin{tabular}{@{}lccc@{}}
\toprule
\textbf{Task} & \textbf{Stage 1 reached} & \textbf{Stage 2 reached} & \textbf{Stage 3 reached} \\ \midrule
Withheld potion pair & 100\% & 100\% & $\sim$7\% \\
Composition 2-hop & 100\% & 100\% & $\sim$30\% \\
Composition 3-hop & 100\% & 100\% & $\sim$48\% \\
Composition 4-hop & 100\% & $\sim$98\% & $\sim$70\% \\
Composition 5-hop & 100\% & 100\% & $\sim$30\% \\ \addlinespace
Decomposition 2-hop & 100\% & 100\% & 100\% \\
Decomposition 3-hop & 100\% & $\sim$72\% & $\sim$44\% \\
Decomposition 4-hop & 100\% & $\sim$88\% & $\sim$30\% \\
Decomposition 5-hop & 100\% & $\sim$37\% & $\sim$13\% \\ \bottomrule
\end{tabular}
\caption{Stage success rate for withheld potion pair / composition / decomposition tasks.}
\label{tab:stage_success_rates_per_task_hop_length}
\end{table}

\begin{figure}[!htb]
    \centering
    \includegraphics[width=\linewidth]{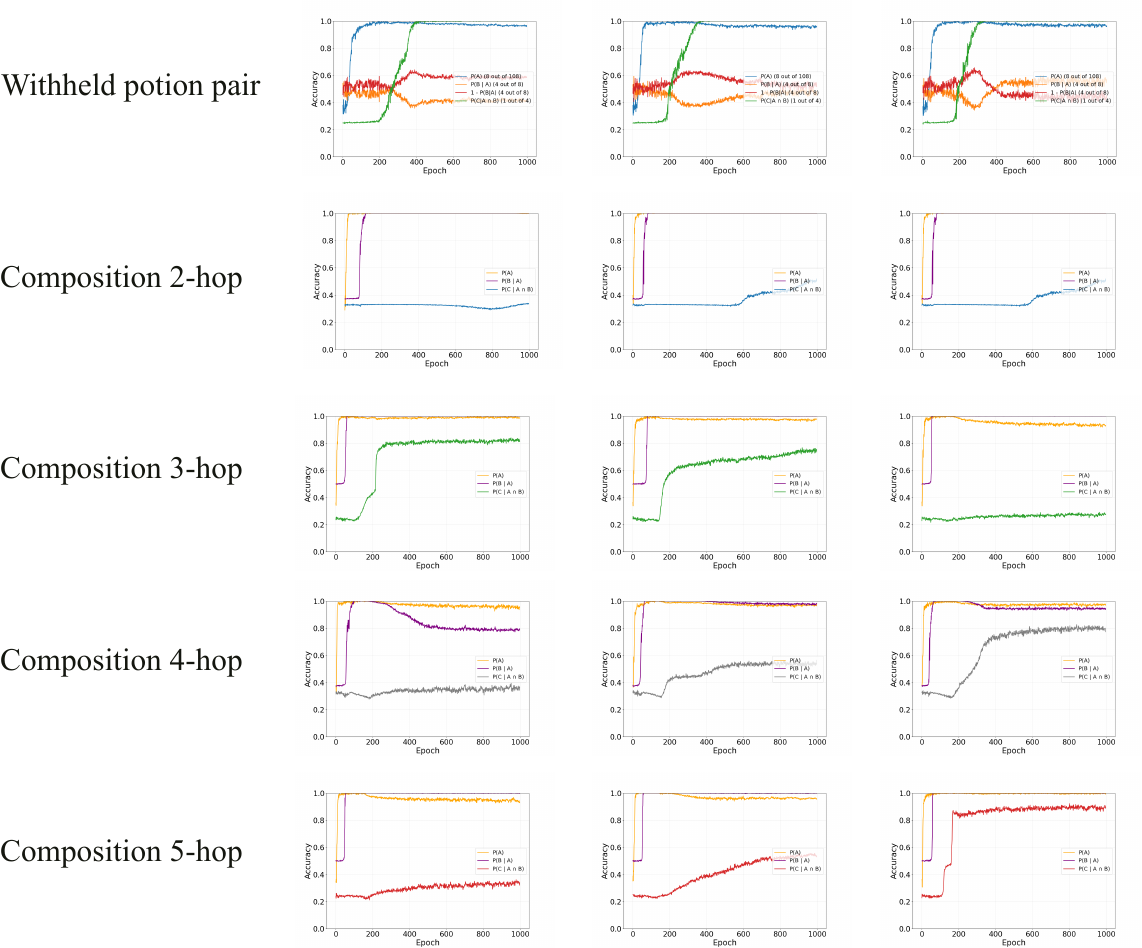}
    \caption{Examples of failure run (non-convergence) stages for the withheld potion and the composition (all hops) tasks. Non-convergence stems from the failure of the final stage to reach 100\% accuracy.}
    \label{fig:failure_run_stage_plots_example_withheld_potion_composition}
\end{figure}

\begin{figure}[!htb]
    \centering
    \includegraphics[width=\linewidth]{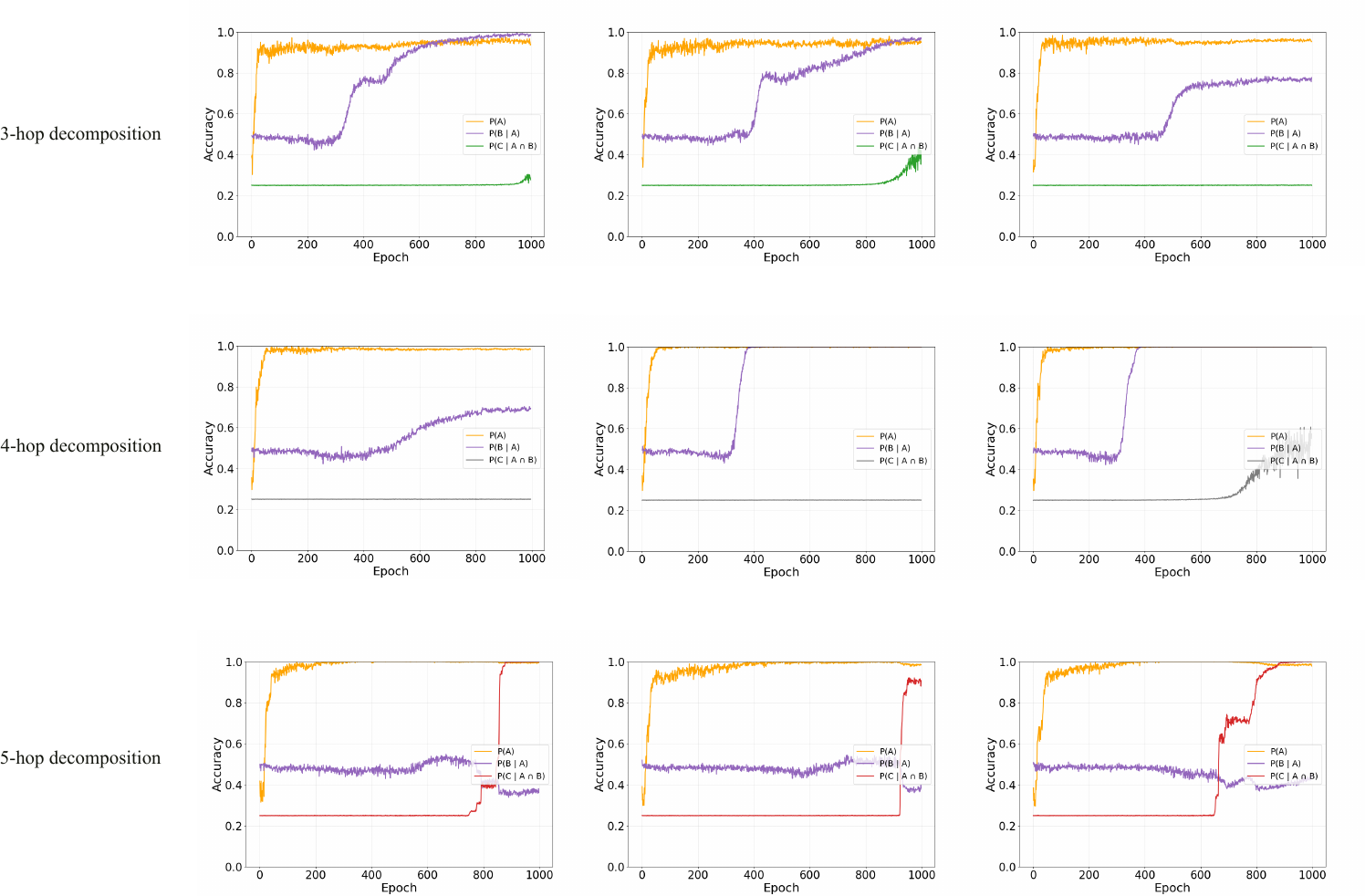}
    \caption{Examples of failure run (non-convergence) stages for the decomposition task, hops 3,4,5. Non-convergence for hops 3,4 stems from the final stage not being able to converge. However, there were cases for the 5-hop decomposition where the order of the non-convergent stages flipped.}
    \label{fig:failure_run_stage_plots_example_decomposition_3_4_5_hops}
\end{figure}

While informative, note that Table \ref{tab:stage_success_rates_per_task_hop_length} does not provide insights into the stage ordering for the failure runs. To this end, we provide the stages for representative failure runs for the three experiments: withheld-potion pair, composition, and decomposition. We show representative failure run stages in Figure \ref{fig:failure_run_stage_plots_example_withheld_potion_composition} for the withheld potion and composition (all hops) tasks. We show the stages for representative failure runs for the decomposition task hops 3,4,5 in Figure \ref{fig:failure_run_stage_plots_example_decomposition_3_4_5_hops}; there were no failure runs for the decomposition 2-hop task.

Note that for the failure runs for all tasks and hop lengths, the model still exhibits stage dynamics, where the in-support stage is reliably learned the earliest. With respect to stage ordering, for the withheld potion and composition tasks, non-convergence of the final stage resulted in overall non-convergence (within the 1000 epoch compute budget).
For the decomposition task hops 3 and 4, overall non-convergence resulted from the final stage not converging within the compute budget. Taken together, for the withheld potion, composition all hops, and decomposition hops 3 and 4, the models exhibit a coarse-to-fine learning of latent structures, where failure to converge stems from the non-convergence of the final stage.

However, 5-hop decomposition failures show reversed stage ordering — stage 3 ($P[C | A \cap B]$) is learned before stage 2 ($P[B | A]$). We hypothesize that for some hyperparameters, the model possibly becomes trapped in a local minimum by focusing on the reward latent structure. To further investigate the flipped stage ordering behavior, and provide evidence of the model focusing on the reward latent structure, we conducted additional diagnostic experiments. \par
\begin{figure}[!htb]
    \centering
    \includegraphics[width=\linewidth]{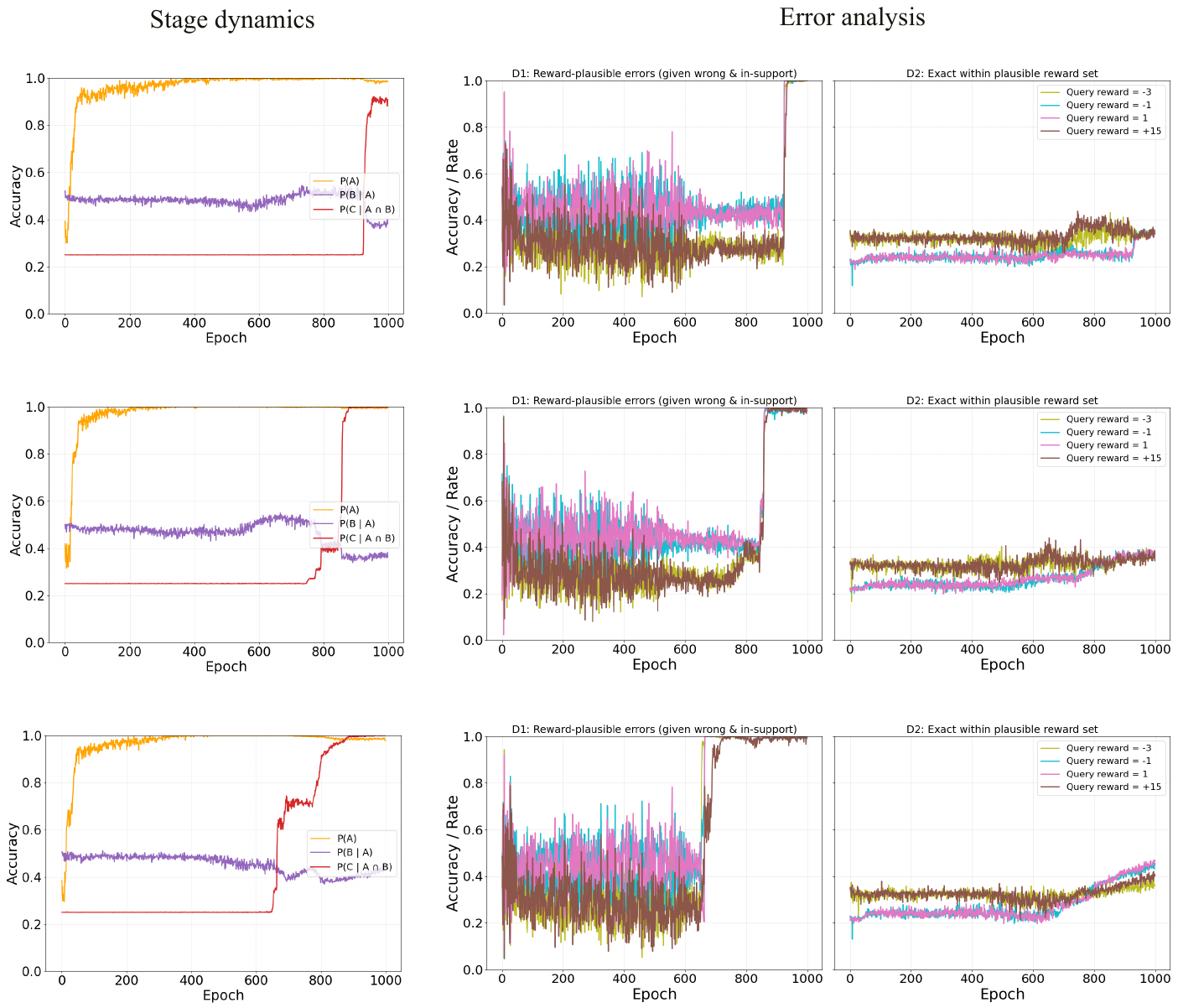}
    \caption{Investigating the 5-hop decomposition flipped stage ordering (left panel) for the failure runs. We used two diagnostic metrics: \textbf{D1}, reward plausible error rate (middle panel) --- the frequency of incorrect predictions with rewards adjacent to the query, and \textbf{D2}, exact match given prediction is in the reward plausible set (right panel). D1 reaching ~100\% indicates that the model is able to restrict its predictions to the plausible rewards for a given query, but is unable to learn D2 (the exact target) within the compute budget, explaining the reversed stage order.}
    \label{fig:decomposition_5_hop_d1_d2_reward_bias_diagnostic}
\end{figure}

Note that the reward distribution follows a consistent topological pattern (Figure 
\ref{fig:example_of_a_chemistry_graph_composition_decomposition}), and target stones only have certain reward values (e.g., +15 is always adjacent to +1). Thus, we used two diagnostic tools focusing on the model's mistakes: \textbf{D1} (reward plausible error rate --- the frequency of incorrect predictions with rewards adjacent to the query), \textbf{D2} (exact match given prediction is in the reward plausible set). Grouping queries by reward feature (similar to Appendix \ref{appendix:adjacency_and_non_support_analysis}), we show the D1 and D2 metrics for the 5-hop decomposition task in Figure \ref{fig:decomposition_5_hop_d1_d2_reward_bias_diagnostic}. D1 reaches 100\%, but D2 remains close to chance, i.e., the model restricts predictions to plausible rewards but fails to learn the exact target, explaining the reversed stage order. Thus, while the decomposition stage factorization broadly holds with the main findings, 5-hop failures show altered staged behavior driven by reward adjacency (Appendix \ref{appendix:adjacency_and_non_support_analysis}). 

\section{Plasticity Windows for Staged Learning}
\label{appendix:additional_plasticity_window_held_out_composition_decomposition}
\label{sec:plasticity_window}

A key question arising from staged learning is whether later stage acquisition depends on continued plasticity in specific layers, and if so, for how long. We address this question by investigating layer-specific \textit{plasticity windows}. Using targeted interventions that freeze individual layers at specific epochs while keeping the rest of the model trainable, we test whether continued plasticity in a layer is causally required for subsequent stage acquisition.

\begin{figure*}[!htb]
    \centering
    \begin{subfigure}[t]{0.49\linewidth}
        \centering
        \includegraphics[width=1.0\linewidth]{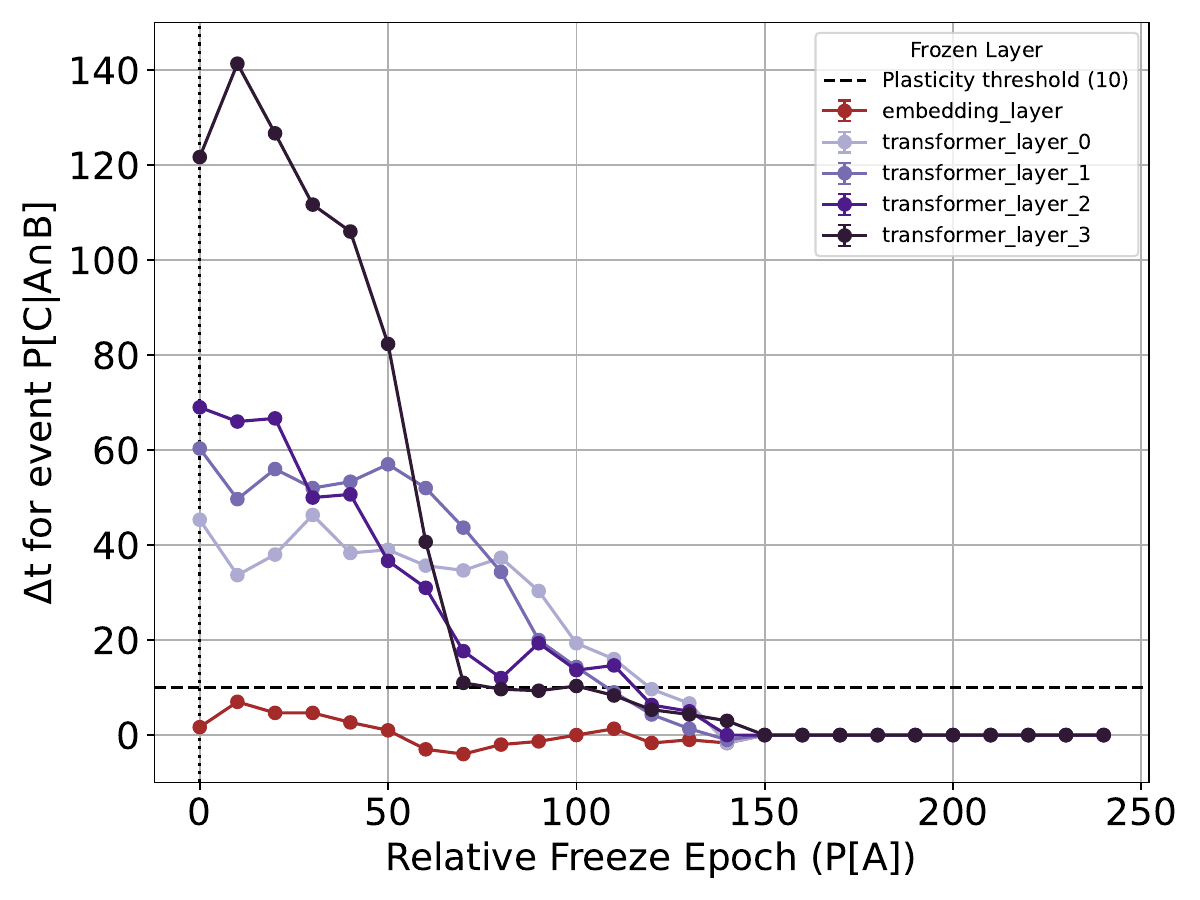}
        \caption{$\Delta t$ for $\mathbb{P}[C \mid A \cap B]$}
        \label{fig:held_out_intermediate_stage_layer_freezing}
    \end{subfigure}
    \hfill
    \begin{subfigure}[t]{0.49\linewidth}
        \centering
        \includegraphics[width=1.0\linewidth]{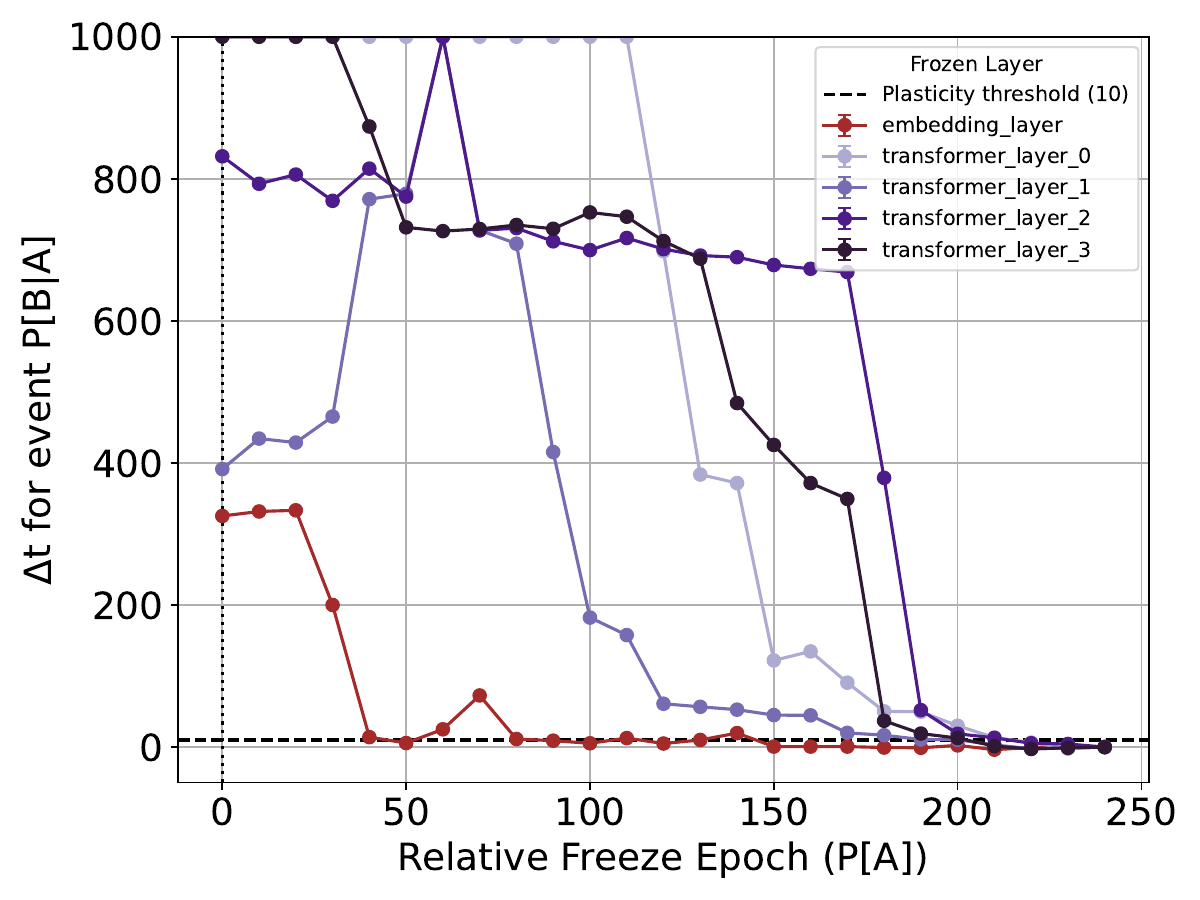}
        \caption{$\Delta t$ for $\mathbb{P}[B \mid A]$}
        \label{fig:held_out_terminal_stage_layer_freezing}
    \end{subfigure}
    \caption{
    Plasticity windows under layer-freezing. Mean $\Delta t$ for acquiring (a) $\mathbb{P}[C \mid A \cap B]$, and (b) $\mathbb{P}[B \mid A]$ across relative freeze epochs $\widetilde{t}_f$ (anchored to $\mathbb{P}[A]$ completion). The dashed line denotes tolerance $\epsilon = 10$. Large delays from early freezing indicate layer-specific plasticity windows, which are notably much shorter for the embedding layer compared to the transformer layers. 
    }
    \label{fig:held_out_layer_freezing_results_representative_seed_both_events}
\end{figure*}

\textbf{Intervention and metric:}
For a layer $\ell$ and freeze epoch $t_f$, we train normally until $t_f$, after which we freeze the parameters of $\ell$ and keep all other parameters trainable under identical hyperparameters. We apply this intervention to each transformer layer and to the token embedding layer. For a stage $v$, we define stage completion $t_v$ as the earliest epoch at which the accuracy $a_v(t) \ge \tau$, with $\tau = 0.95$, and remains above it for a patience window of $P=3$ epochs: $t_v = \min\{t : a_v(t') \ge \tau,\ \forall t' \in [t, t+P]\}$. We measure the intervention effect relative to a control (unfrozen) run as $\Delta t_v(\ell,t_f) = t_v^{\text{freeze}(\ell,t_f)} - t_v^{\text{control}}$, and set $\Delta t_v$ to a maximum value of 1000 if the model fails to converge. We account for negligible variation in delay with a tolerance $\epsilon = 10$. Finally, we express freeze times relative to control stage completion: $\widetilde{t_f} = t_f - t_{\mathbb{P}[A]}^{\text{control}}$ to account for different model initializations. We anchor all results on the completion of the first stage $\mathbb{P}[A]$ (in-support) and conduct the freezing interventions every 10 epochs. We provide the results for the withheld potion task (Section \ref{sec:latent_structure_learning}) in
Figure \ref{fig:held_out_layer_freezing_results_representative_seed_both_events}, which shows the mean $\Delta t$ for both $\mathbb{P}[C| A \cap B]$ (a), and $\mathbb{P}[B | A]$ (b).

\par
For both events, freezing transformer layers after $\mathbb{P}[A]$ substantially delays subsequent stage acquisition, and $\Delta t$ decreases with increasing $\widetilde{t}_f$. This demonstrates that continued plasticity is required for several epochs beyond learning $\mathbb{P}[A]$. The stage $\mathbb{P}[B|A]$ is more sensitive to early freezing of multiple transformer layers, often resulting in large delays or non-convergence within the compute budget (1000 epochs). In contrast, freezing early causes shorter but noticeable delays for $\mathbb{P}[C | A \cap B]$.

Compared to transformer layers, freezing the embedding layer causes no noticeable delay ($\Delta t \le \epsilon$) for $\mathbb{P}[C | A \cap B]$ with limited impact on $\mathbb{P}[B | A]$. Overall, these interventions reveal layer-specific \textit{plasticity windows} during which continued adaptation of a layer is required for subsequent stage acquisition, and freezing after this window produces no noticeable convergence delays.
\par

\begin{figure}[!htb]
    \centering
    \includegraphics[width=0.8\linewidth]{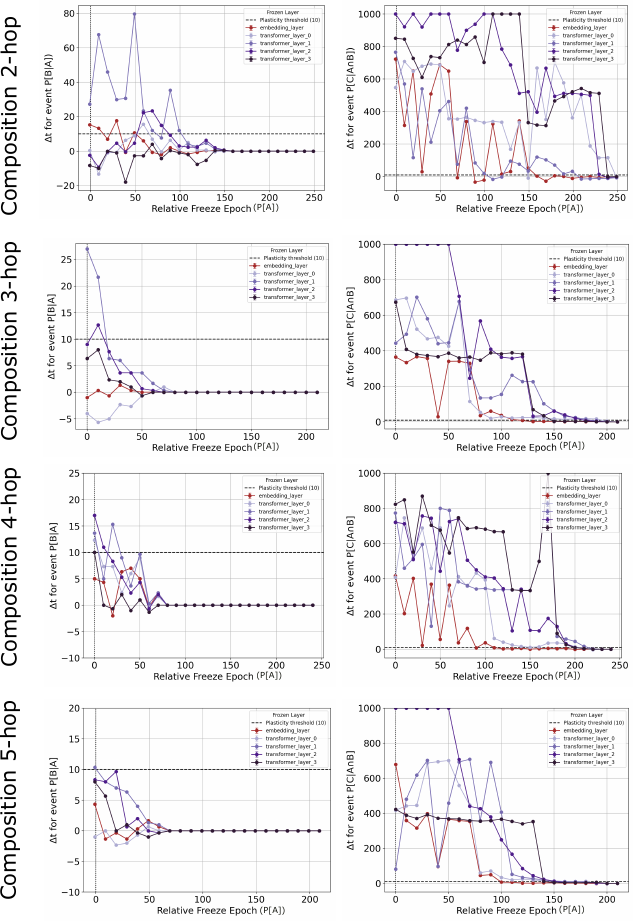}
    \caption{Plasticity windows for composition all hops under layer-freezing. Mean $\Delta t$ for acquiring $\mathbb{P}[B \mid A]$ (left column), and $P[C \mid A \cap B]$ (right column) across relative freeze epochs $\widetilde{t}_f$ (anchored to $\mathbb{P}[A]$ completion). The dashed line denotes tolerance $\epsilon = 10$. Large delays from early freezing indicate layer-specific plasticity windows, with the embedding layer often showing shorter windows compared to many transformer layers.} 
    \label{fig:composition_plasticity_windows_all_hops}
\end{figure}

\begin{figure}[!htb]
    \centering
    \includegraphics[width=0.8\linewidth]{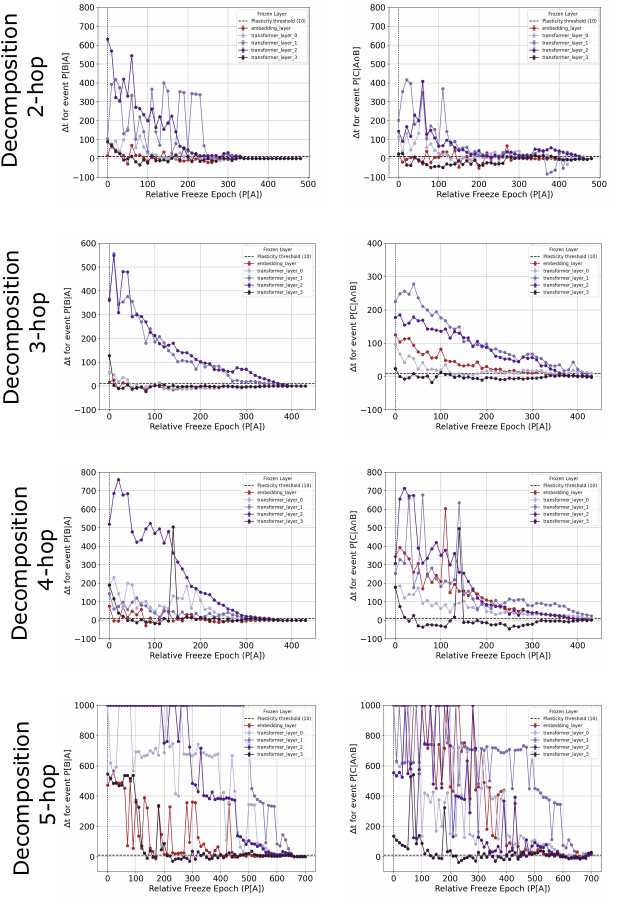}
    \caption{Plasticity windows for decomposition all hops under layer-freezing. Mean $\Delta t$ for acquiring $\mathbb{P}[B \mid A]$ (left column), and $P[C \mid A \cap B]$ (right column) across relative freeze epochs $\widetilde{t}_f$ (anchored to $\mathbb{P}[A]$ completion). The dashed line denotes tolerance $\epsilon = 10$. Large delays from early freezing indicate layer-specific plasticity windows, with the embedding layer often showing shorter windows compared to many transformer layers.}
    \label{fig:decomposition_plasticity_windows_all_hops}
\end{figure}

We provide the plasticity window results for composition in Figure \ref{fig:composition_plasticity_windows_all_hops}, and decomposition in Figure \ref{fig:decomposition_plasticity_windows_all_hops}. 
These figures extend the layer-freezing analysis to the composition and decomposition settings and across hop lengths ($hl \in \{2,3,4,5\}$). 
Across both settings, freezing transformer layers causes noticeable delays, indicating that continued adaptation in these layers is important for timely subsequent stage acquisition. Compared to the embedding layer, transformer layers often exhibit longer plasticity windows, though this pattern is not uniform across all hops and stages (e.g., $\mathbb{P}[C| A \cap B]$ for 3-hop decomposition in Figure \ref{fig:decomposition_plasticity_windows_all_hops}). These results provide additional insights into layer-specific plasticity windows for composition and decomposition tasks.

\clearpage
\end{document}